\documentclass[twoside,11pt]{article}

\usepackage[abbrvbib,preprint]{jmlr2e}

\usepackage{dsfont}
\usepackage{booktabs}
\usepackage{subcaption}
\usepackage{xcolor}         % colors
\usepackage{psfrag}
\usepackage[cmex10]{amsmath}
\usepackage{amsfonts,latexsym,hyperref}
\usepackage[captionskip=0pt]{floatrow} % for putting figure and table side-by-side
\usepackage{algorithm,algpseudocode} % algorithmic

 \renewcommand{\hat}{\widehat}
 \renewcommand{\bar}{\overline}
 \newcommand{\Real}{{\mathbb{R}}}

 \newcommand{\tran}{^{\text{\textsf{T}}}}

 \renewcommand{\vec}[1]{\ensuremath{\boldsymbol{#1}}}
 \newcommand{\hvec}[1]{\ensuremath{\hat{\boldsymbol{#1}}}}

 \newcommand{\defn}{\triangleq}

 \DeclareMathOperator{\E}{E}

 \DeclareMathOperator*{\argmin}{arg\,min}
 \DeclareMathOperator*{\argmax}{arg\,max}

\usepackage{grffile}
\usepackage{booktabs}

\graphicspath{{./figures_send/}}

 \newcommand{\putTable}[3]{\begin{table}[t!]
                            \centering
                            \caption{#2}
                            \label{tab:#1}
                            #3
                          \end{table} }
 \newcommand{\capFrag}[2]{}
 \newcommand{\capTable}[2]{}

 \newcommand{\mc}[1]{\ensuremath{\mathcal{#1}}}
 \newcommand{\norm}[1]{\ensuremath{\| #1 \|}}

 \renewcommand{\eqref}[1]{(\ref{eq:#1})}
 
 \newcommand{\Figref}[1]{Figure~\ref{fig:#1}}
 \newcommand{\figref}[1]{Figure~\ref{fig:#1}}
 \newcommand{\tabref}[1]{Table~\ref{tab:#1}}
 \newcommand{\secref}[1]{Section~\ref{sec:#1}}
 
 \newcommand{\appref}[1]{Appendix~\ref{app:#1}}

 \renewcommand{\algref}[1]{Algorithm~\ref{alg:#1}}
 
 \newcommand{\scoreglrt}{t_{\mathsf{GLRT}}}
 \newcommand{\scoreeigk}{t_{\mathsf{eig},k}}
 
 \DeclareMathOperator{\pbeta}{beta}

 \newcommand{\scorecosj}{s_{\mathsf{cos},j}}
 \newcommand{\scorecosone}{s_{\mathsf{cos},1}}

 \newcommand{\scorenormj}{s_{\mathsf{norm},j}}
 \newcommand{\scorenormone}{s_{\mathsf{norm},1}}

 \newcommand{\scoreshiftj}{s_{\mathsf{shift},j}}
 \newcommand{\scoreshiftone}{s_{\mathsf{shift},1}}

 \newcommand{\lamcon}{\lambda^{\mathsf{con}}}
 \newcommand{\lamshift}{\lambda^{\mathsf{shift}}}
 \newcommand{\scorecsi}{s_{\mathsf{CSI}}}
 \newcommand{\scoresupcsi}{s_{\mathsf{supCSI}}}
 \newcommand{\scoresupcsij}{s_{\mathsf{supCSI},j}}
 \newcommand{\scoresupcsione}{s_{\mathsf{supCSI},1}}

 \newcommand{\scoresupocsvmj}{s_{\mathsf{supOCSVM},j}}
 \newcommand{\scoresupocsvmone}{s_{\mathsf{supOCSVM},1}}

 \newcommand{\scoresupmahj}{s_{\mathsf{supMah},j}}
 \newcommand{\scoresupmahone}{s_{\mathsf{supMah},1}}

 \DeclareMathOperator{\cossim}{sim}
 \DeclareMathOperator{\softmax}{softmax}
 \DeclareMathOperator{\GLR}{GLR}
 \DeclareMathOperator{\LR}{LR}
 
 \DeclareMathOperator{\betadist}{beta}
 \DeclareMathOperator{\invcdf}{inv\,cdf}

 \newcommand{\term}[1]{\emph{#1}}
 \newcommand{\inlierPx}{P_\mathsf{x}}
 
 \newcommand{\so}{s} % observed score when computing p-value
 \newcommand{\qt}{\mathsf{q}}
 \newcommand{\train}{_\mathsf{train}}
 \newcommand{\traink}{_{\mathsf{train},k}}
 \newcommand{\test}{_\mathsf{test}}
 \newcommand{\val}{_\mathsf{val}}
 \newcommand{\bon}{_{\mathsf{bon}}}
 \newcommand{\fish}{_{\mathsf{fisher}}}
 \newcommand{\simes}{_{\mathsf{simes}}}
 \newcommand{\BH}{_{\mathsf{BH}}}
 \newcommand{\stouf}{_{\mathsf{stouffer}}}
 \newcommand{\cifar}{^{\mathsf{cifar10}}}
 \newcommand{\svhn}{^{\mathsf{svhn}}}

 \newcommand{\citeo}[1]{} % suppresses the citation

 \newcommand{\qthresh}{\mathsf{a}}

\usepackage{lastpage}
\firstpageno{1}

\begin{document}
\setlength{\arraycolsep}{0.5mm}

\title{Score Combining for Contrastive OOD Detection}

\author{%
\name Edward T. Reehorst
\email edward.reehorst@us.af.mil\\
\addr Sensors Directorate \\
    Air Force Research Lab\\
    Wright-Patterson Air Force Base, OH 45433-7320, USA
\AND
\name Philip Schniter
\email schniter.1@osu.edu\\
\addr Department of Electrical and Computer Engineering\\
    The Ohio State University\\
    Columbus, OH 43210, USA%
}

\editor{My editor}

\maketitle

\begin{abstract}%
In out-of-distribution (OOD) detection, one is asked to classify whether a test sample comes from a known inlier distribution or not.
We focus on the case where the inlier distribution is defined by a training dataset and there exists no additional knowledge about the novelties that one is likely to encounter. 
This problem is also referred to as 
novelty detection, 
one-class classification,
and unsupervised anomaly detection. 
The current literature suggests that contrastive learning techniques are state-of-the-art for OOD detection.
We aim to improve on those techniques by combining/ensembling their scores using the framework of null hypothesis testing and, in particular, a novel generalized likelihood ratio test (GLRT).
We demonstrate that our proposed GLRT-based technique outperforms the state-of-the-art CSI and SupCSI techniques from \citep{Tack:NIPS:20} in dataset-vs-dataset experiments with CIFAR-10, SVHN, LSUN, ImageNet, and CIFAR-100, as well as leave-one-class-out experiments with CIFAR-10.
We also demonstrate that our GLRT outperforms the score-combining methods of Fisher, Bonferroni, Simes, Benjamini-Hochwald, and Stouffer in our application.
\end{abstract}


\section{Introduction}
Out-of-distribution (OOD) detection refers to the problem of classifying whether a test sample comes from a known \term{inlier} distribution or not.
%Unlike supervised binary classification, one has no examples of non-inlier samples.
OOD detection has wide-ranging applications in, e.g., 
intrusion detection in cybersecurity; 
fraud detection in finance, insurance, healthcare, and telecommunications;
industrial fault and damage detection;
monitoring of infrastructure and stock markets;
acoustic novelty detection;
medical diagnosis and disease outbreak detection;
even detection in the earth sciences;
and scientific discovery in chemistry, bioinformatics, genetics, physics, and astronomy---%
see the recent comprehensive overview \citep{Ruff:PROC:21} for specific citations.
%radar \citep{Chakravarthy:RAD:20}

We focus on the case where the inlier distribution is defined by a dataset $\mc{X}=\{\vec{x}_i\}$ and there is no prior knowledge about non-inlier samples.
In the literature, this problem is also referred to as 
unsupervised anomaly detection, 
%out-of-distribution detection,
novelty detection,
one-class classification,
and open-set classification
\citep{Ruff:PROC:21}.  
Although semi-supervised variations of OOD detection have been proposed using, e.g., outlier exposure \citep{Hendrycks:ICLR:19,Ruff:ICLR:20} 
and transfer learning with pre-trained networks \citep{Reiss:CVPR:21,Deecke:ICML:21,Reiss:AAAI:23}, %could add \citep{Xue:23} model zoo, but 0 other citations 
%they are not compatible with all applications and hence 
we focus on the purely unsupervised setting.

Many methods have been developed for OOD detection.
Among classical approaches, there are those based on density estimation, e.g., 
using
Mahalanobis distance \citep{Laurikkala:IDAMP:00}, %\citep{Gnanadesikan:IBS:72}
Gaussian mixture models (GMMs) \citep{Bishop:IEEV:94},
nearest-neighbor distance \citep{Ramaswamy:ICMD:00,Breunig:ICMD:00}, %\citep{Hautamaki:ICPR:04}
or 
kernel density estimation (KDE) \citep{Latecki:MLDM:07}. %\citep{Parzen:AMS:62}.
However, the density estimation is fundamentally difficult with high-dimensional data, such as images, due to the curse of dimensionality \citep{Zimek:SADM:12}.
There also exist classical approaches based on
kernel PCA \citep{Hoffmann:PR:07}
and support vector machines (SVMs) \citep{Scholkopf:NC:01,Tax:ML:04},
but selecting kernels and hand-crafting relevant features is challenging.
For example, many contemporary OOD detection tasks aim to distinguish high-level semantic differences (e.g., objects from non-normal classes) over low-level sensory differences (e.g., texture defects) \citep{Ahmed:AAAI:20}.
In the raw feature space, semantic novelties can be very close to inliers according to traditional $\ell_p$ and kernel-based distances.
For example, an image of a dog with fur that is cat-like fur texture and color similar to a cat can be more similar in raw pixel space than various cat breeds among themselves.

To address the shortcomings of these classical approaches, a wide variety of deep neural network (DNN)-based OOD detection techniques have been proposed.
As described in recent surveys \citep{Chalapathy:19,DiMattia:19,Pang:CS:21,Ruff:PROC:21},
they include methods based on deep autoencoders, 
deep SVMs, 
deep generative networks, such as generative adversarial networks and normalizing flows, 
and self-supervised methods.
Among these, self-supervised methods like the contrastive shifted instances (CSI) \citep{Tack:NIPS:20} have been particularly effective, perhaps because self-supervised learning seems to exhibit inductive biases towards semantic representations \citep{Ahmed:AAAI:20}.

Most OOD detection techniques construct an \term{inlier score} that quantifies how likely a given sample is to be an inlier.
A test sample is classified as an inlier if its score exceeds a predetermined threshold.
In some cases, several intermediate scores are computed and then combined to form the final score.
For example, with self-supervised techniques, it's common to compute intermediate scores using pretext tasks, augmented views, and/or distribution shifts \citep{Golan:NIPS:18,Hendrycks:NIPS:19,Bergman:ICLR:20,Tack:NIPS:20,Georgescu:CVPR:21,Khalid:CVPRW:22}.
However, the score combining methodologies in these works are heuristic.
Our contributions are as follows.
\begin{enumerate}
\item 
We propose to use the framework of null-hypothesis testing to design principled score-combining techniques for self-supervised OOD detection.
\item 
We propose a novel score-combining technique based on a generalized likelihood ratio test (GLRT), and we demonstrate that it performs better than the classical Fisher, Bonferroni, Simes/Benjamini-Hochwald (BH), Stouffer, and ALR tests in our application.
\item
We show how to achieve false-alarm-rate guarantees with a finite-sample validation set.
\item
We demonstrate improvements over the state-of-the-art CSI and SupCSI \citep{Tack:NIPS:20} contrastive OOD detectors on dataset-vs-dataset experiments with CIFAR-10, SVHN, LSUN, ImageNet, and CIFAR-100 and on leave-one-class-out experiments with CIFAR-10.
\end{enumerate}

To our knowledge, this is the first work to apply the null-hypothesis testing framework to self-supervised OOD detection.
Null-hypothesis testing has been applied to other forms of OOD detection, though.
For example, 
\citet{Bergamin:AISTATS:22} used Fisher's method combine two scores produced by deep generative models.
\citet{Kaur:AAAI:22} aggregated conformity scores measuring transformational equivariance into a single p-value, which was thresholded for OOD detection.
\citet{Haroush:ICLR:22} computed p-values for each class, layer, and channel in a DNN trained for supervised classification, and combined those scores into a final p-value using a combination of Fisher's and Simes' methods.
\citet{Magesh:22} used the BH procedure \citep{Benjamini:JRSSb:95} to aggregate scores from the feature maps of a DNN trained for supervised classification in a way that yields guarantees on the false-alarm rate.
Note that, in \citep{Haroush:ICLR:22} and \citep{Magesh:22}, the goal was to determine whether a \emph{given} trained DNN behaves differently on a test sample compared to the samples used to train it.
However, depending on how it was trained, that DNN could throw away information that is useful for OOD detection. 
Our approach is different:
Given an inlier dataset $\mc{X}$, we aim to design/train a DNN to detect whether a test sample is OOD or not.

\section{Background}

\subsection{Self-supervised learning for OOD detection} \label{sec:self}

There are two main approaches to self-supervised learning \citep{Ericsson:SPM:22}.  %\citep{Albelwi:Entropy:22}
The first is to design a pretext task that yields a set of labels and then use those labels to train a DNN in a supervised manner.
An example is training a DNN to predict the rotation of an image by either 0, 90, 180, or 270 degrees \citep{Gidaris:ICLR:18}. 
The intermediate features of the trained DNN can then be used for other tasks, such as OOD detection \citep{Golan:NIPS:18,Hendrycks:NIPS:19,Bergman:ICLR:20}.
When the dimensionality of these features is sufficiently small, classical approaches to OOD detection are effective.

The other main approach to self-supervised learning is contrastive learning, which trains the DNN to distinguish between semantically similar augmented views of a given data sample and other data samples.
Concretely, let $A_0(\cdot)$ and $A_1(\cdot)$ denote two augmentations randomly selected from an augmentation set $\mc{A}$ (e.g., for images, crop, horizontal flip, color jitter, and grayscale, as in \citep{Tack:NIPS:20}), 
%and let $\vec{x}_i^a\defn A_a(\vec{x}_i)$ for $a\in\{0,1\}$ be the corresponding augmentations of training sample $\vec{x}_i$.
%Then let $\vec{f}_i^a \defn \vec{f}_{\vec{\theta}}(\vec{x}_i^a)$ for $a\in\{0,1\}$ be the corresponding outputs 
and let $g_{\vec{\theta}}(A_a(\vec{x}_i))$ for $a\in\{0,1\}$ be the corresponding augmented outputs from DNN $g_{\vec{\theta}}(\cdot)$. 
Using the shorthand $\vec{g}_i^a\defn g_{\vec{\theta}}(A_a(\vec{x}_i))$,
and denoting cosine similarity by 
\begin{align}
\cossim(\vec{g}_i,\vec{g}_{i'})\defn \frac{\vec{g}_i\tran\vec{g}_{i'}}{\norm{\vec{g}_i}\norm{\vec{g}_{i'}}},
\end{align}
contrastive learning trains $\vec{\theta}$ to maximize $\cossim(\vec{g}_i^0,\vec{g}_i^1)$ while minimizing $\cossim(\vec{g}_i^a,\vec{g}_{i'}^{a'})$ for $(i',a')\neq(i,a)$. 
For example, SimCLR \citep{Chen:ICML:20} trains $\vec{\theta}$ to minimize $\mc{L}_{\text{con}}(\vec{\theta})$ over a $B$-sample batch, where
\begin{align}
\mc{L}_{\text{con}}(\vec{\theta})
&= 
\frac{1}{2B}\sum_{i=1}^B \sum_{a=0}^1
-\ln \frac{\exp(\cossim(g_{\vec{\theta}}(\vec{x}_i^a),g_{\vec{\theta}}(\vec{x}_i^{1-a}))/\tau)}
{\sum_{(i',a')\neq (i,a)}\exp(\cossim(g_{\vec{\theta}}(\vec{x}_i^a),g_{\vec{\theta}}(\vec{x}_{i'}^{a'}))/\tau)}
\label{eq:simclr}
\end{align}
and $\tau>0$ is a tunable temperature.
Once trained, the outputs and/or intermediate-layer features of $g_{\vec{\theta}}(\cdot)$ can be employed for OOD detection using classical methods.
For example, \citet{Winkens:20} used Mahalanobis distance while \citet{Sohn:ICLR:21} used KDE and one-class SVM (OC-SVM).

SimCLR drives the inlier features $\{\vec{g}_i^a\}$ towards a uniform distribution on the hypersphere \citep{Wang:ICML:20}, which makes it difficult to distinguish inliers from OOD samples. 
Thus, \citet{Tack:NIPS:20} and \citet{Sohn:ICLR:21} proposed to augment the data $\{\vec{x}_i\}$ using \term{distribution-shifting transformations} $T_j(\cdot)\in\mc{T}$, such as $\{0^\circ, 90^\circ,180^\circ,270^\circ\}$ image rotations. 

The contrastive shifted instances (CSI) technique from \citep{Tack:NIPS:20} achieves state-of-the-art performance by training with a loss that combines SimCLR with distribution-shift prediction.
Once the DNN $g_{\vec{\theta}}(\cdot)$ is trained, CSI evaluates and combines $3|\mc{T}|=12$ different scores for the above shift set $\mc{T}$. 
Some scores are computed from the outputs of $g_{\vec{\theta}}(\cdot)$ and others from the logits $w(\vec{x})\in\Real^{|\mc{T}|}$ of a linear shift-classifier head that is attached 
at $f_{\vec{\theta}}(\cdot)$, which is 
two layers behind the output of $g_{\vec{\theta}}(\cdot)$. 
CSI uses three base scores:
\begin{align}
\scorecosj(\vec{x})
&\defn \max_{\vec{x}'\in\mc{X}\train} \cossim\big(g_{\vec{\theta}}(R_j(\vec{x})),g_{\vec{\theta}}(R_j(\vec{x}'))\big) 
\label{eq:cosj}\\
\scorenormj(\vec{x})
&\defn \big\|g_{\vec{\theta}}(R_j(\vec{x}))\big\|  
\label{eq:normj}\\
\scoreshiftj(\vec{x})
&\defn w_j(R_j(\vec{x}))
\label{eq:shiftj},
\end{align}
defined for shifts $j\in\{1,\dots,|\mc{T}|\}$, and then combines them into a single score as follows:
\begin{align}
\scorecsi(\vec{x})
&\defn \sum_{j=1}^{|\mc{T}|} \Big[ \lamcon_{j} \scorecosj(\vec{x})\scorenormj(\vec{x}) + \lamshift_{j} \scoreshiftj(\vec{x})\Big]
\label{eq:csi}\\
\lamcon_{j} 
&\defn \left[\frac{1}{n}\sum_{i=1}^n \scorenormj(\vec{x}_i) \right]^{-1}
\text{~~and~~}
\lamshift_{j} 
\defn \left[\frac{1}{n}\sum_{i=1}^n \scoreshiftj(\vec{x}_i) \right]^{-1} 
\label{eq:lambda_csi} .
\end{align}

When the inlier dataset $\{\vec{x}_i\}$ has class labels $y_i\in\{1,\dots,K\}$, SupCSI \citep{Tack:NIPS:20} exploits them to outperform CSI.
SupCSI trains with SupCLR \citep{Khosla:NIPS:20} instead of SimCLR, as well as joint class/shift prediction.  
Once the DNN is trained, the class/shift logits $\vec{\ell}^{K\times |\mc{T}|}(\cdot)$ are used to construct the score
\begin{align}
\scoresupcsi(\vec{x})
&= \max_{k=1,\dots,K} \left[\softmax\left(\frac{1}{|\mc{T}|}\sum_{j=1}^{|\mc{T}|} \vec{\ell}_{:,j}(R_j(\vec{x}))\right)\right]_k 
\label{eq:supcsi} ,
\end{align}
where $\vec{\ell}_{:,j}(\cdot)$ is the $j$th column of $\vec{\ell}(\cdot)$ and
$[\softmax(\vec{\ell})]_k \defn \exp(\ell_k)/\sum_{k'=1}^K \exp (\ell_{k'})$.
Although CSI and SupCSI yield state-of-the-art performance, their score-combining rules \eqref{csi}-\eqref{supcsi} are heuristic, leaving room for improvement.
Also, it's not clear how additional scores could be incorporated into \eqref{csi}-\eqref{supcsi}.

\subsection{Statistical hypothesis testing for OOD detection} \label{sec:hypo}

\textbf{Null hypothesis testing.} 
Consider the problem of choosing between two hypotheses, $H_0$ (null) and $H_1$ (alternative), where 
$H_0$ says that $\vec{x}$ was drawn from some distribution $\inlierPx$ and $H_1$ says that it was not drawn from $\inlierPx$.
Applied to OOD detection, $\vec{x}$ is an inlier under $H_0$ and an OOD sample under $H_1$.
Given a test statistic $s(\vec{x})$ that measures confidence in $H_0$ (i.e., an \term{inlier score}), we reject the null hypothesis when $s(\vec{x})\leq \tau$ for some threshold $\tau$.
%This is known as a left-tailed test.
A false alarm, or \term{type-1} error, occurs when we mistakenly reject an $\vec{x}$ generated under $H_0$. 
The probability of a false alarm, $\alpha=\Pr\{s(\vec{X})\leq \tau \,|\, H_0\}$, is known as the \term{significance level}.  
Usually, $\alpha$ is determined by the application and $\tau$ is adjusted accordingly. 

Often, \term{p-values} are used to quantify the significance of a test outcome.
For an observed $\so=s(\vec{x})$, the p-value is defined as follows, where $\vec{X}$ denotes a random version of $\vec{x}$:
\begin{align}
q(\so)\defn\Pr\{s(\vec{X})\leq \so\,|\,H_0\}
\label{eq:pvalue}.
\end{align}
The p-value indicates the probability of observing a test outcome that is equal or more extreme than $\so$ given that $H_0$ is true.
If we treat $\so$ as a random variable (i.e., $S\defn s(\vec{X})$ for random test sample $\vec{X}$), and assume that the cumulative distribution function (cdf) of $S$ under $H_0$ is continuous, then $q(S)$ is uniformly distributed on $[0,1]$ under the null hypothesis \citep{Wasserman:Book:04}, i.e., $\Pr\{q(S)\leq \alpha\,|\,H_0\}=\alpha$ for any $\alpha\in(0,1)$. 
So, given an observed $\so$, to reject the null hypothesis at a significance of $\alpha$, we simply check whether $q(\so)\leq\alpha$.
%For a given significance level $\alpha$, we'd like to choose a statistic $s(\cdot)$ that maximizes $\Pr\{s(\vec{x})\leq \tau \,|\, H_1\}$, known as the \term{power} of the test.

\textbf{Global null tests.} 
Suppose that we have access to $m>1$ statistics $\{s_l(\vec{x})\}_{l=1}^m$ of the form described above, and that we associate them with the respective hypothesis pairs $\{H_{0l},H_{1l}\}_{l=1}^m$. 
The \term{global null hypothesis} refers to the case where all null hypotheses are true, i.e., $H_0 = \cap_{l=1}^m H_{0l}$.
Because the statistics $\{s_l(\vec{x})\}$ can differ significantly over $l$, it's much easier to combine them after they have been converted to p-values $\vec{q}=[q_1,\dots,q_m]\tran$.

Famous methods for combining p-values include 
Fisher's method \citep{Fisher:Book:92}, 
Bonferroni's method \citep{Wasserman:Book:04}, 
Sime's method \citep{Simes:BIO:86},
and the BH procedure \citep{Benjamini:JRSSb:95}.  
Each method is associated with a test statistic $t(\vec{q})$ that rejects the global null $H_0$ when $t(\vec{q})\leq \tau$, for an appropriately chosen threshold $\tau$.
These statistics can be written as 
\begin{align}
t\fish(\vec{q}) &\defn \sum_{l=1}^m \ln q_l\\
t\bon(\vec{q}) &\defn \min_l q_l\\
t\simes(\vec{q})=t\BH(\vec{q}) &\defn \min_l q_{(l)}/l, 
\end{align}
where $q_{(l)}$ is the $l$th sorted p-value such that $q_{(1)}\leq q_{(2)} \leq \cdots \leq q_{(m)}$.

\section{Proposed method} \label{sec:proposed}

Suppose we have $m$ inlier score functions $\{s_l(\cdot)\}_{l=1}^m$, where $s_l:\Real^d\rightarrow \Real$ and a larger value of $s_l(\vec{x})$ indicates increasing confidence in hypothesis $H_0$, i.e., that $\vec{x}$ is an inlier.
As described in \secref{self}, such scores commonly arise in self-supervised OOD detectors. 
Our goal is to combine these $m$ scores $\{s_l(\vec{x})\}$ into a single test statistic $t(\vec{x})$ that we can threshold via $t(\vec{x})\leq\tau$ to classify whether $\vec{x}$ is an inlier or not.

%\subsection{Classical p-value combining methods}

\begin{figure}[t]
    \centering
    \newcommand{\sz}{30mm}
    \newcommand{\szz}{29mm}
    \includegraphics[height=\sz,trim=12 10 10 13,clip]{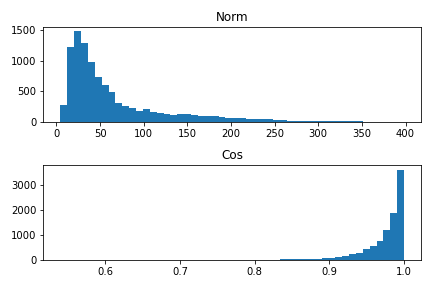}
    \hfill
    \includegraphics[height=\sz,trim= 0 0 0 4,clip]{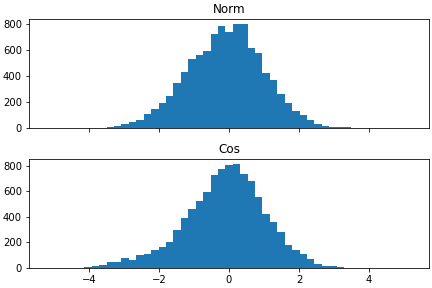}
    \hfill
    \includegraphics[height=\szz,trim=30 18 55 30,clip]{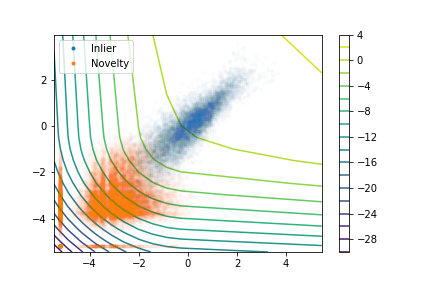}
    \mbox{}
    \caption{Left: Histogram of CSI's ``norm'' score $\{s_1(\vec{x}_i)\}_{i=1}^n$ (top) and ``cos'' score $\{s_2(\vec{x}_i)\}_{i=1}^n$ (bottom) for CIFAR-10 inliers $\mc{X}\test\cifar$.
    Center: Histograms of the corresponding empirical z-values $\{\hat{z}_1(s_1(\vec{x}_i))\}_{i=1}^n$ and $\{\hat{z}_2(s_2(\vec{x}_i))\}_{i=1}^n$ with $\hat{z}_1(\cdot)$ and $\hat{z}_2(\cdot)$ constructed using CIFAR-10 inliers $\mc{X}\train\cifar$.
    Right: Scatter plot of $\hat{z}_1(s_1(\vec{x}_i))$ (horizontal axis) versus $\hat{z}_2(s_2(\vec{x}_i))$ (vertical axis) for CIFAR-10 inliers $\mc{X}\test\cifar$ and SVHN novelties $\mc{X}\test\svhn$.
Contours show the value of the proposed GLRT score $\scoreglrt(\vec{z})$ versus $z_1$ and $z_2$.}
    \label{fig:score_plots}
    \vspace{2mm}
\end{figure}

One of the main challenges in combining scores is that they can have very different distributions (see the left pane of \figref{score_plots}).
If the scores were simply summed together, the larger scores could overwhelm the smaller ones, regardless of whether they were more informative or not.
As discussed in \secref{hypo}, a common way to circumvent this problem is to transform each score $s_l(\vec{x})$ to its p-value $q_l(s_l(\vec{x}))$ and combine the p-values using, e.g., Bonferroni, Fisher, or Simes/BH.
%We numerically evaluate these combining approaches in \secref{experiments}.

%\subsection{z-values}

Another principled approach to combining scores uses \term{z-values}, 
\begin{align}
z_l(s) \defn \Phi^{-1}(q_l(s)) \text{~for p-value $q_l(s)$ and the standard-normal cdf $\Phi(\cdot)$}
\label{eq:zvalue},
\end{align}
which are standard-normal distributed under $H_0$ rather than uniform; see \citep{Cousins:07} for a historical perspective;
Because $s_l(\vec{x})$ is an inlier score and $z_l(s)$ is monotonically increasing in $s$, we can also interpret $z_l(s_l(\vec{x}))$ as an inlier score.
A well-known way to combine z-values $\vec{z}=[z_1,\dots,z_m]\tran$ is to use \term{Stouffer's statistic}, 
\begin{align}
t\stouf(\vec{z})\defn \frac{1}{\sqrt{m}}\sum_{l=1}^m z_l,
\end{align}
which we evaluate in \secref{experiments}.
%talk about weighted Stouffer?
%https://onlinelibrary.wiley.com/doi/abs/10.1111/j.1420-9101.2005.00917.x
Below, we propose a different way to combine z-values.

In the sequel, when dealing with random $\vec{X}$, we will abbreviate $z_l(s_l(\vec{X}))$ as $Z_l$, which is itself random, and define $\vec{Z}\defn[Z_1,\dots,Z_m]\tran$.
Likewise, for a deterministic observed test sample $\vec{x}$, we will abbreviate $z_l(s_l(\vec{x}))$ as $z_l$, which is itself deterministic, and define $\vec{z}\defn[z_1,\dots,z_m]\tran$.

\subsection{The negative-means problem and a GLRT-based solution} \label{sec:glrt}

We formulate OOD detection as the following hypothesis testing problem:
\begin{subequations}
\label{eq:NMP}
\begin{align}
&H_0:~\vec{Z} \sim \mc{N}(\vec{0},\vec{I}) 
\label{eq:NMP0} \\
&H_1:~\vec{Z} \sim \mc{N}(\vec{\mu},\vec{I}) \text{~with~} \mu_l\leq-\epsilon \text{~for all~}l=1,\dots,m,
\label{eq:NMP1} 
\end{align}
\end{subequations}
which we will refer as the \term{negative-means} (NM) problem.
Here, the constant $\epsilon> 0$ is specified in advance but $\vec{\mu}=[\mu_1,\dots,\mu_m]\tran$ is unknown.
The motivation for \eqref{NMP0} is that, by construction, each $Z_l$ is standard-normal under $H_0$.
The motivation for \eqref{NMP1} is that, because each $z_l$ constitutes an inlier score, the (unknown) mean of $Z_l$ under $H_1$ should be $\epsilon$ less than the mean of $Z_l$ under $H_0$, for some $\epsilon> 0$.
To be clear, \eqref{NMP} models the z-values as independent, even though we don't expect this property to hold in practice.
We discuss alternative formulations that circumvent this independence assumption in \appref{alternatives}.

If $\vec{\mu}$ in \eqref{NMP} was known, we could solve \eqref{NMP} using a likelihood ratio test (LRT) \citep{Lehmann:Book:05}, i.e.,
\begin{align}
\LR(\vec{z}) \underset{H_1}{\overset{H_0}{\gtrless}} \tau'
\quad\text{for}\quad
\LR(\vec{z}) \defn \frac{p(\vec{z}|H_0)}{p(\vec{z}|H_1)}
\label{eq:LR} ,
\end{align}
where $p(\vec{z}|H_i)$ denotes the probability density function (pdf) of $\vec{Z}$ under $H_i$.
The LRT is known to be Neyman-Pearson optimal, meaning that the probability of detection is maximized for a fixed false-alarm rate \citep{Lehmann:Book:05}.
Thresholding $\LR(\vec{z})$ in \eqref{LR} would be equivalent to thresholding the log-LR
\begin{align}
\ln \LR(\vec{z})
&= -\tfrac{1}{2}\|\vec{z}\|^2 + \tfrac{1}{2}\|\vec{z}-\vec{\mu}\|^2
= -\vec{\mu}\tran\vec{z} + \tfrac{1}{2}\|\vec{\mu}\|^2
\label{eq:logLR},
\end{align}
and thus equivalent to thresholding $-\vec{\mu}\tran\vec{z}$ at some $\tau$.

In \eqref{NMP}, however, $\vec{\mu}$ is unknown, and thus \eqref{NMP} is a \term{composite} hypothesis testing problem \citep{Lehmann:Book:05}.
For such problems, a \term{uniformly most powerful} (UMP) test exists when there exists a Neyman-Pearson test statistic that is invariant to the unknown parameters \citep{Lehmann:Book:05}.
In our case, because the test statistic $-\vec{\mu}\tran\vec{z}$ depends on the direction of the unknown $\vec{\mu}$, a UMP does not exist unless $m=1$. 
When $m=1$, because $-\mu_1\geq\epsilon>0$, we can form a UMP test by rejecting $H_0$ whenever $z_1\leq\tau$.
%Whenever $m > 1$, a UMP test does not exist because the test statistic \eqref{log_lr} changes with the direction of the unknown, $\vec{\mu}$.

% \subsection{GLRT}

A common approach to composite hypothesis testing is the generalized LRT (GLRT) \citep{Lehmann:Book:05}. 
For the NM problem \eqref{NMP}, the GLRT takes the form
\begin{align}
\GLR(\vec{z}) \underset{H_1}{\overset{H_0}{\gtrless}} \tau'
\quad\text{for}\quad
\GLR(\vec{z}) \defn \frac{p(\vec{z}|H_0)}{\max_{\vec{\mu}: \mu_l \leq -\epsilon~\forall l} p(\vec{z}|H_1, \vec{\mu})}
\label{eq:glrt_def} .
\end{align}
\citet{Wei:AS:19} studied \eqref{glrt_def} when $\epsilon=0$, in which case the set $\{\vec{\mu}:\mu_l\leq -\epsilon~\forall l\}$ is a closed convex cone.
For that case, they proved that the GLRT is optimal in a minimax sense.
We focus on the case that $\epsilon>0$, which tends to improve performance in our application, as shown in \secref{experiments}.

To simplify the GLRT \eqref{glrt_def}, we first derive the denominator-maximizing value of $\vec{\mu}$ as
\begin{eqnarray}
%\vec{\mu}^* &\defn& 
\argmax_{\vec{\mu}:\mu_l\leq-\epsilon~\forall l} p(\vec{z}|H_1,\vec{\mu})
= \argmin_{\vec{\mu}:\mu_l\leq-\epsilon~\forall l} \|\vec{z}-\vec{\mu}\|^2
%~~\Rightarrow~~ \mu_l^* = z_l^-
= \vec{z}^-
~~\text{for}~~
[\vec{z}^-]_l = z_l^- 
\defn \min\{z_l,-\epsilon\}
%\begin{cases} z_l & \text{if~}z_l\leq -\epsilon\\ -\epsilon & \text{else}. \end{cases}
\quad
\label{eq:z_minus_simple} .
\end{eqnarray}
Then we plug this value of $\vec{\mu}$ into the log-LR expression \eqref{logLR} to obtain the log-GLR 
\begin{align}
%\scoreglrt(\vec{z})
\ln \GLR(\vec{z})
&= -(\vec{z}^-)\tran\vec{z} + \tfrac{1}{2}\|\vec{z}^-\|^2
%= (\tfrac{1}{2}\vec{\mu}^* -\vec{z})\tran\vec{\mu}^* 
= (\tfrac{1}{2}\vec{z}^- -\vec{z})\tran\vec{z}^- 
\defn \scoreglrt(\vec{z})
\label{eq:logGLR},
\end{align}
which we use as our proposed test statistic. 
In other words, we reject the null hypothesis $H_0$ whenever $\scoreglrt(\vec{z})\leq\tau$ for some appropriately chosen threshold $\tau$.

%\appref{glrt_plots} show plots of $\scoreglrt(z)$ versus $z\in\Real$ for comparison with Fisher's and Stouffer's test statistics.
%\subsection{Plots of the GLRT test statistic} \label{app:glrt_plots}

\begin{figure}[t]
    \centering
    \newcommand{\sz}{40mm}
    \hfill
    \includegraphics[height=\sz,trim=35 5 40 15,clip]{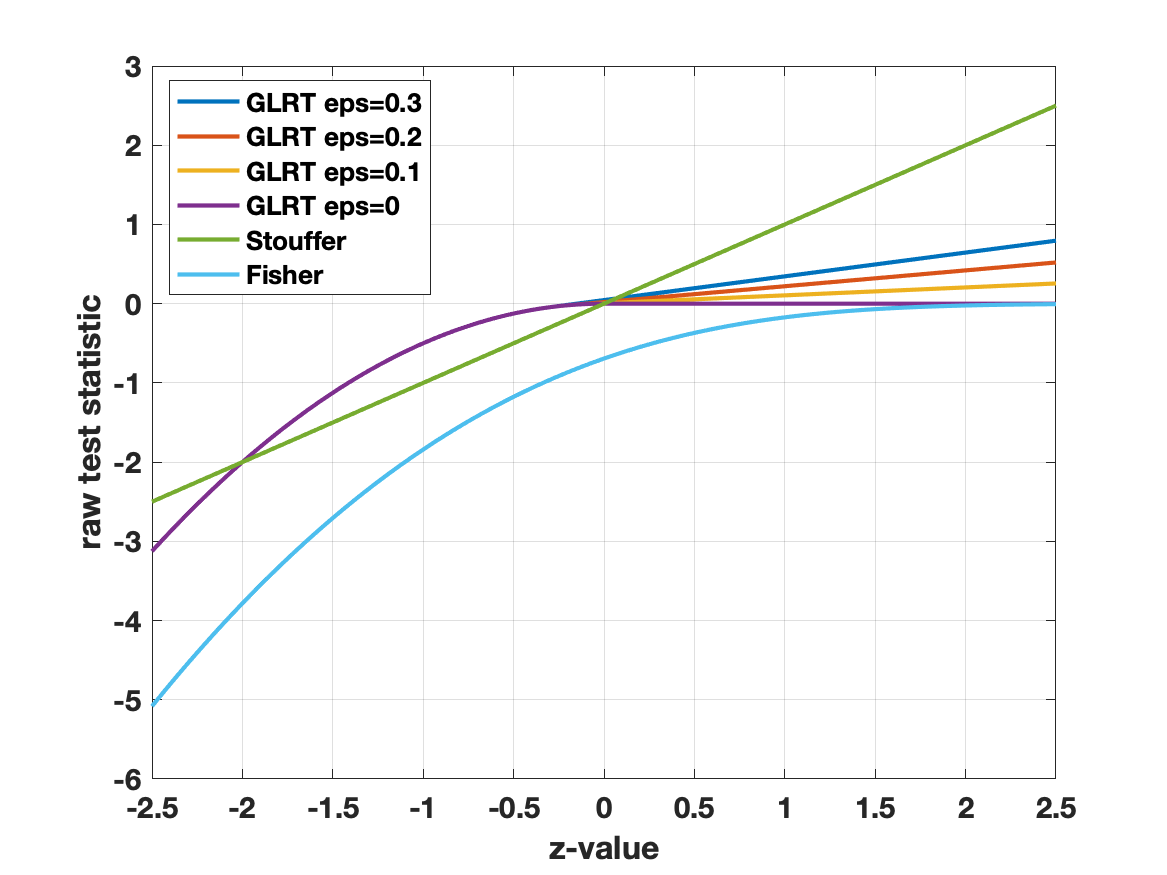}
    \hfill
    \includegraphics[height=\sz,trim=35 5 40 15,clip]{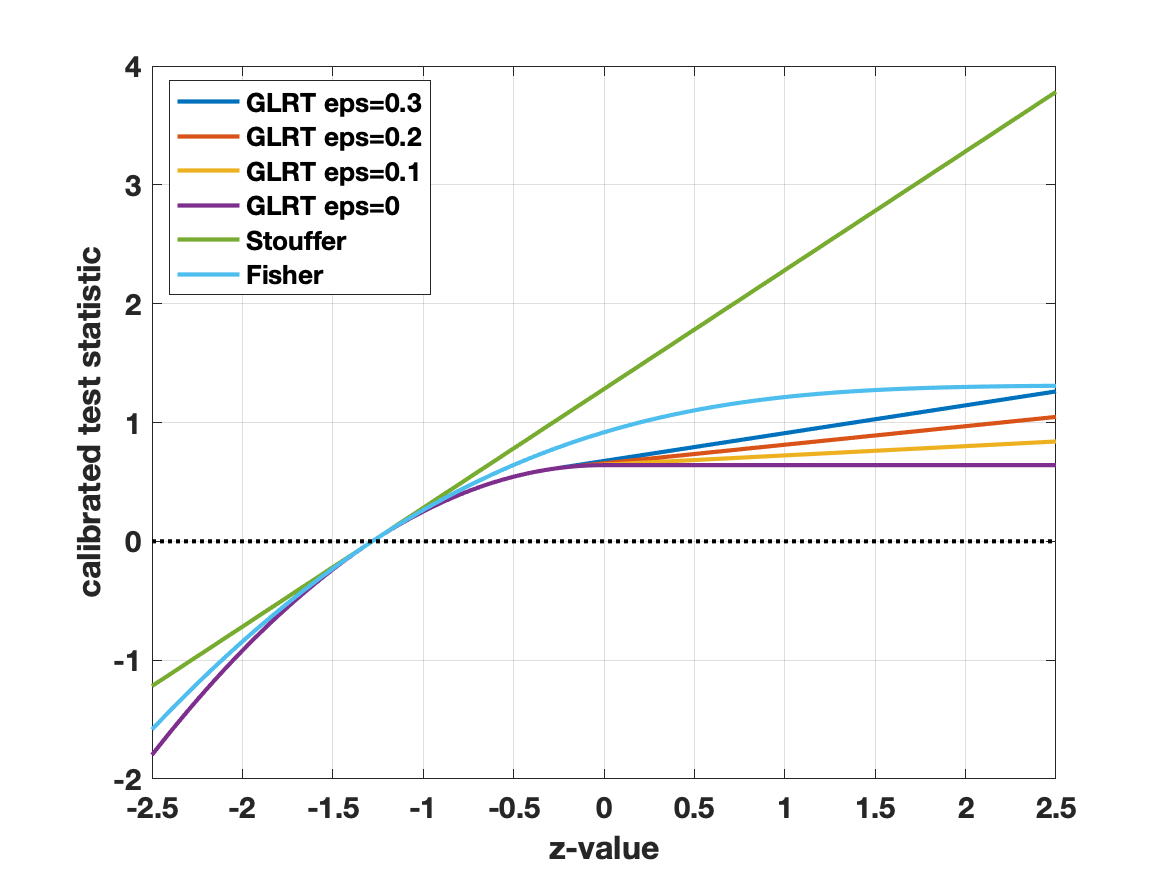}
    \hfill
    \mbox{}
    \caption{Left:
    Plot of $\scoreglrt(z)$ versus $z\in\Real$ for 
    $\epsilon\in\{0,0.1,0.2,0.3\}$, along with $t\stouf(z)=z$ and $t\fish(z)=\ln\Phi(z)$.
    Right: a plot of the same test statistics 
    after shifting and scaling $t(z)$ so that 
    $\Pr\{t(z)\leq 0\,|\,H_0\}=0.1$ and $t'(z_0)=1$ for $z_0$ such that $t(z_0)=0$.}
    %\vspace{-2mm}
    \label{fig:glrt_plots}
\end{figure}

In \figref{glrt_plots}(a), we plot $\scoreglrt(z)$ versus $z\in\Real$ for $\epsilon\in\{0,0.1,0.2,0.3\}$, for comparison to $t\stouf(z)=z$ and $t\fish(z)=\ln\Phi(z)$.
To facilitate an easier visual comparison, we exploit the fact that the test $t(z)\leq \tau$ is equivalent to $\bar{t}(z) \defn C(t(z)-\tau)\leq 0$ for any constant $C>0$.
We then choose $(C,\tau)$ to calibrate each test so that $\Pr\{\bar{t}(z)\leq 0\,|\,H_0\}=0.1$ (i.e., each test has a false-alarm rate of $0.1$) and $\bar{t}'(z_0)=1$ for $z_0$ such that $\bar{t}(z_0)=0$ (i.e., each test has unit slope at the false-alarm threshold).
We plot the calibrated tests in \figref{glrt_plots}(b).

\subsection{Empirical z-values}

In practice, it's not possible to compute the z-value in \eqref{zvalue} because we don't know the cdf of $s_l(\vec{X})$ under $H_0$. 
Thus we propose to use \term{empirical z-values}, which are computed from training data $\mc{X}\train=\{\vec{x}_i\}_{i=1}^n$ via
\begin{align} 
\hat{z}_l(s) \defn \Phi^{-1} (\hat{F}_l(s)), \quad
\hat{F}_l(s) \defn \frac{1}{n}\sum_{i=1}^n \mathds{1}\{s_l (\vec{x}_i) \leq s\},
\end{align}
where $\mathds{1}(\cdot)$ is the indicator function and 
$\hat{F}_l(\cdot)$ is the empirical cdf of $\{s_l(\vec{x}_i)\}_{i=1}^n$.
See the middle plot of \figref{score_plots} for an example.

For the case of a given threshold $\tau$, we summarize the proposed OOD detector in \algref{glrt_tau} using a streamlined notation that writes the z-values $\hat{z}_l(s(\vec{x}))$ and GLRT score $\scoreglrt(\vec{z}(\vec{x}))$ directly as functions of $\vec{x}$.

It is not immediately obvious, though, how to choose the threshold $\tau$ when the goal is to achieve a target false-alarm rate of $\alpha$.
Doing so is the subject of the next section.

\begin{algorithm}[t]
\caption{The proposed GLRT-based OOD detector for a given threshold $\tau$}
\label{alg:glrt_tau}
\begin{algorithmic}[1]
\Require
test sample $\vec{x}$,
test threshold $\tau$,
training samples $\{\vec{x}_i\}_{i=1}^n$,
%training set $\mc{X}\train$,
scores $\{s_l(\cdot)\}_{l=1}^m$,
NM parameter $\epsilon>0$.
\State
$\forall l=1,\dots,m: 
\hat{F}_l(\vec{x}) = \frac{1}{n}\sum_{i=1}^n \mathds{1}\{s_l (\vec{x}_i) \leq s_l(\vec{x})\}
%\text{~where~} \{\vec{x}_i\}_{i=1}^n = \mc{X}\train 
$
\State
$\forall l=1,\dots,m: \hat{z}_l(\vec{x}) = \Phi^{-1}(\hat{F}_l(\vec{x})) 
$
\State
$\scoreglrt(\vec{x}) = \sum_{l=1}^m \big(\frac{1}{2}\hat{z}_l^-(\vec{x}) -\hat{z}_l(\vec{x})\big)\hat{z}_l^-(\vec{x})$ for $\hat{z}_l^-(\vec{x})=\min\{\hat{z}_l(\vec{x}),-\epsilon\}$
\State
\Return ``OOD'' if $\scoreglrt(\vec{x})\leq \tau$, else ``inlier.''
\end{algorithmic}
\end{algorithm}

\subsection{False-alarm-rate guarantees} \label{sec:conformal}
%\subsection{Details on false-alarm rate guarantees} \label{app:conformal}

In this section we consider how to guarantee a target false-alarm rate of $\alpha$ for a statistical test $t(\cdot)$ that rejects $H_0$ when $t(\vec{x})\leq \tau$ for some threshold $\tau$.
Ideally, we would like to choose $\tau$ such that the false-alarm rate $\Pr\{t(\vec{X})\leq \tau\,|\,H_0\}$ is upper bounded by a user-specified $\alpha\in(0,1)$.
If we knew the cdf of $t(\vec{X})$ under $H_0$, then we could compute the p-value $\qt(t)\defn\Pr\{t(\vec{X})\leq t\,|\,H_0\}$, which is essentially a way of transforming the random variable $t(\vec{X})$ to a new random variable $\qt(t(\vec{X}))$ 
that is uniform on $[0,1]$ under $H_0$. 
In this case, the test $\qt(t(\vec{x})) \leq \alpha$ would yield a false-alarm rate of $\alpha$.

When the cdf of $t(\vec{X})$ under $H_0$ is unknown, however, the above procedure cannot be used.
An alternative is to replace the p-value $\qt(t)$ by the \term{conformal p-value}
\begin{align}
\hat{\qt}(t) \defn \frac{1+|\{\vec{x}_i\in\mc{X}\val: t(\vec{x}_i)\leq t\}|}{1+v} ,
\end{align}
which is computed using a size-$v$ validation set $\mc{X}\val$ drawn i.i.d.\ under $H_0$ and independent of $\vec{x}$ and any data used to design $t(\cdot)$.
In this case,
%Here, $t$ represents the test statistic $t(\vec{x})$ for a given test sample $\vec{x}$.
%If we treat $\vec{X}$ as random, then $T\defn t(\vec{X})$ becomes random, and 
it's known \citep{Vovk:ACML:12} % Bates:AS:23
that 
\begin{align}
\E[ \Pr\{\hat{\qt}(t(\vec{X}))\leq \qthresh\,|\,H_0,\mc{X}\val\} ] \leq \qthresh \text{~for all~} \qthresh\in(0,1), 
\end{align}
where the expectation is taken over $\mc{X}\val$.
Thus, when averaged over \emph{many} validation sets $\mc{X}\val$, the test ``$\hat{\qt}(t(\vec{x}))\leq\alpha$'' provides a false-alarm rate of $\alpha$.
But in practice we have only \emph{one} validation set, and so this average-case result does not apply.
%Rather, we must acknowledge that our validation set $\mc{X}\val$ may be an unlucky choice.

Our approach is to choose a more conservative threshold ``$\qthresh$'' on $\hat{\qt}(t(\vec{x}))$ such that, with probability at least $1-\delta$, a false-alarm rate of at most $\alpha$ will be attained.
To do this, we treat $\mc{X}\val$ as random, in which case the false-alarm rate $\Pr\{\hat{\qt}(t(\vec{X}))\leq \qthresh\,|\,H_0,\mc{X}\val\}$ is itself random and distributed as \citep{Vovk:ACML:12} % Bates:AS:23
\begin{align}
\Pr\{\hat{\qt}(t(\vec{X}))\leq \qthresh\,|\,H_0,\mc{X}\val\}
&\sim
\betadist\big(\lfloor (v\!+\!1)\qthresh\rfloor,v\!+\!1\!-\!\lfloor (v\!+\!1)\qthresh\rfloor\big)
\label{eq:beta},
\end{align}
which concentrates at $\qthresh$ as the validation size $v\rightarrow\infty$ \citep{Vovk:ACML:12}. %Bates:AS:23
Given this distribution, we propose to use bisection search to find a value of $\qthresh$ that guarantees $\alpha_{\delta}\leq \alpha$, where $\alpha_{\delta}$ is the false-alarm-rate value that yields a beta upper-tail probability of exactly $\delta$. 
%, with probability at least $1-\delta$, the false-alarm rate will be at most $\alpha$.

\begin{figure}[t]
    \centering
    \hfill
    \begin{subfigure}{0.3\textwidth}
        \includegraphics[width=\textwidth,trim=25 5 40 20,clip]{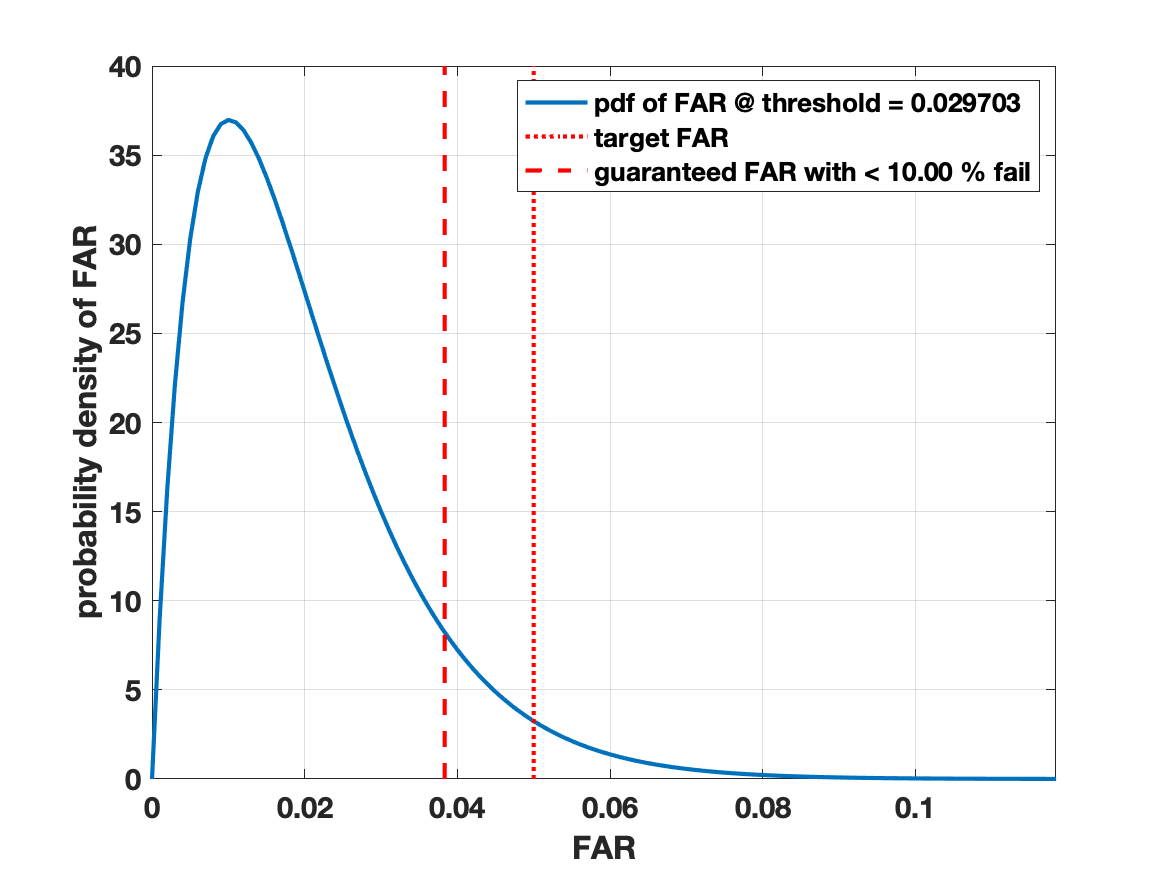}
        \caption{$v=100$}
    \end{subfigure}
    \hfill
    \begin{subfigure}{0.3\textwidth}
        \includegraphics[width=\textwidth,trim=25 5 40 20,clip]{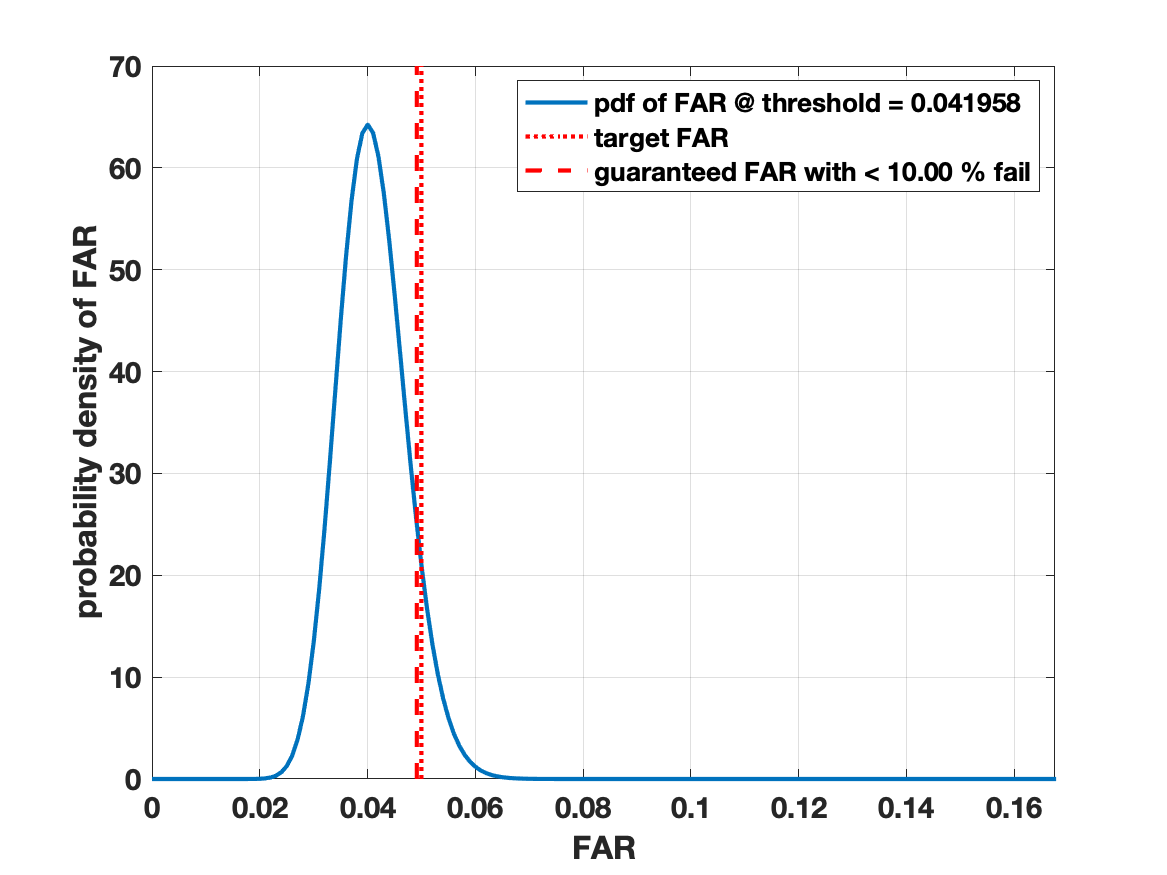}
        \caption{$v=1000$}
    \end{subfigure}
    \hfill
    \begin{subfigure}{0.3\textwidth}
        \includegraphics[width=\textwidth,trim=25 5 40 20,clip]{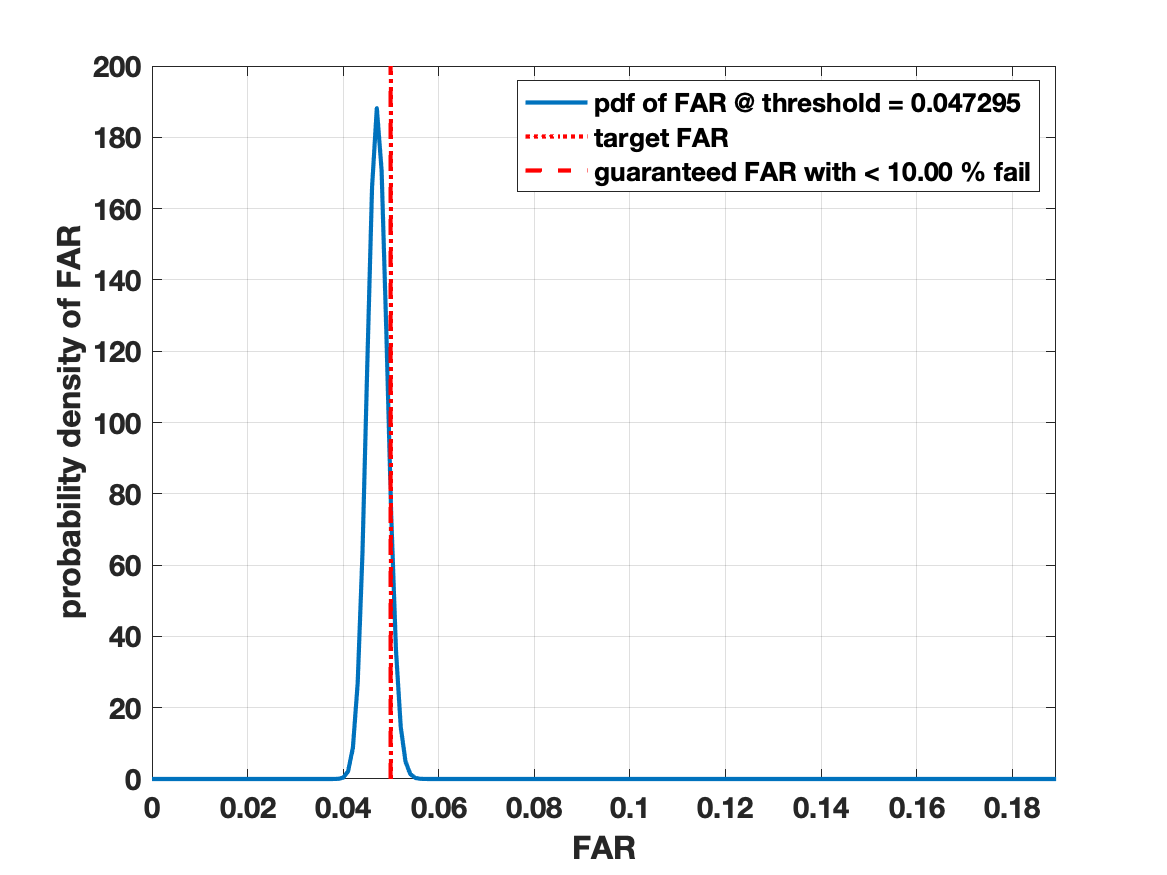}
        \caption{$v=10000$}
    \end{subfigure}
    \hfill
    \mbox{}
    %\vspace{-2mm}
    \caption{Examples of the beta pdf from \eqref{beta} and thresholds $\qthresh$ that guarantee a target false-alarm rate (FAR) of at most $\alpha=0.05$ (red dotted line) with failure probability of at most $\delta=0.1$.  The achieved false alarm rate, $\alpha_{\min}$, is shown by the dashed red line.} 
    \label{fig:conformal}
\end{figure}

A procedure to find this $\qthresh$ is given in \algref{bisection}, but a few comments are in order.
First notice that, due to the floor operations in \eqref{beta}, there exist half-open intervals of $\qthresh$ that yield the same $\betadist$ parameters and thus the same false-alarm rate statistics.
These half-open intervals take the form of $[\frac{l}{v+1},\frac{l+1}{v+1})$ for $l=0,\dots,v$.
Thus it suffices for the bisection algorithm to find a pair $(\qthresh_{\min},\qthresh_{\max})$ such that 
$\qthresh_{\min}\in[\frac{l}{v+1},\frac{l+1}{v+1})$ yields false-alarm rate $\alpha_{\min}<\alpha$ with probability at least $1-\delta$,
and
$\qthresh_{\max}\in[\frac{l+1}{v+1},\frac{l+2}{v+1})$ yields false-alarm rate $\alpha_{\max}>\alpha$ with probability at least $1-\delta$,
for some $l\in\{0,\dots,v-1\}$. 
For that value of $l$, it then suffices to choose any $\qthresh\in[\frac{l}{v+1},\frac{l+1}{v+1})$, and \algref{bisection} chooses a value close to the right boundary of the interval.
The achieved false-alarm rate is then
$\alpha_{\min}\defn\invcdf \big(1-\delta,\betadist(l,v+1-l)\big)$.
\Figref{conformal} shows examples of the $\betadist$ pdf and $\alpha_{\min}$ for validation sizes $v\in\{100,1000,10000\}$.

\begin{algorithm}[t]
\caption{Bisection search to find a conformal-p-value threshold of $\qthresh$ that yields a false-alarm rate of $\alpha$ with probability at least $1-\delta$}
\label{alg:bisection}
\begin{algorithmic}[1]
\Require
validation size $v=|\mc{X}\val|$,
maximum false-alarm rate $\alpha\in(0,1)$,
maximum failure rate $\delta\in(0,1)$.
\State
Initialize $\qthresh_{\min} = 0$, $\qthresh_{\max} = 1$.
\While{$\lfloor(v+1)\qthresh_{\max}\rfloor-\lfloor(v+1)\qthresh_{\min}\rfloor>1$}
    \State $\qthresh = 0.5(\qthresh_{\min}+\qthresh_{\max})$
    \State $l = \lfloor(v+1)\qthresh\rfloor$
    \State $\alpha_{\delta} = \invcdf\big(1-\delta,\betadist(l,v+1-l)\big)$
    \If{$\alpha_{\delta} > \alpha$}
        \State $\qthresh_{\max}=\qthresh$
    \Else
        \State $\qthresh_{\min}=\qthresh$
    \EndIf
\EndWhile
%\State
%\% For any $\qthresh\in[\frac{l}{v+1},\frac{l+1}{v+1})$, have that $\Pr\big[ \Pr\{\hat{\qt}(t(\vec{X}))\leq\qthresh\,|\,H_0,\mc{X}\val\}\leq\alpha \big]\geq 1-\delta$
\State
\Return $\qthresh=\frac{l+0.99}{v+1}$ 
\end{algorithmic}
\end{algorithm}

Finally, \algref{tau} summarizes how conformal p-values can be used for OOD detection white guaranteeing a false-alarm rate of at most $\alpha$ with probability at least $1-\delta$.
%To attain a target false-alarm rate of $\alpha$ with probability at least $1-\delta$, we can use a size-$v$ calibration set $\mc{X}\val$ and conformal p-values as summarized in \algref{tau}.  
%%the statistical analysis from \citep{Bates:AS:23}, 

\begin{algorithm}[t]
\caption{OOD detection achieving a false-alarm rate of at most $\alpha$ with probability at least $1-\delta$}
\label{alg:tau}
\begin{algorithmic}[1]
\Require
test sample $\vec{x}$,
generic inlier test $t(\cdot)$,
%test sample $\vec{x}$,
%training $\mc{X}\train$,
validation $\mc{X}\val$ of size $v$,
%scores $\{s_l(\cdot)\}_{l=1}^m$,
%NMP parameter $\epsilon>0$,
$\alpha\in(0,1)$,
$\delta\in(0,1)$.
\State
Use \algref{bisection} to find $\qthresh\in(0,1)$ such that 
$\Pr\{\pbeta(\lfloor (v\!+\!1)\qthresh\rfloor,v\!+\!1\!-\!\lfloor (v\!+\!1)\qthresh \rfloor) \leq \alpha \} \geq 1-\delta$  
\State
Compute the conformal p-value
$\displaystyle \hat{\qt}(\vec{x}) = \frac{1+|\{\vec{x}_i\in\mc{X}\val: t(\vec{x}_i)\leq t(\vec{x})\}| }{1+v}$ 
\State
\Return ``OOD'' if $\hat{\qt}(\vec{x})\leq \qthresh$, else ``inlier.''
\end{algorithmic}
\end{algorithm}

% THIS IS THE "ASYMPTOTIC" METHOD FROM THE BATES PAPER
%\begin{algorithm}[t]
%\color{red}
%\caption{The proposed GLRT-based OOD detector for $\delta$-guaranteed false-alarm rate $\alpha$}
%\label{alg:glrt_alpha2}
%\begin{algorithmic}[1]
%\Require
%test sample $\vec{x}$,
%training $\mc{X}\train$,
%validation $\mc{X}\val$,
%scores $\{s_l(\cdot)\}_{l=1}^m$,
%NMP parameter $\epsilon>0$,
%false-alarm rate $\alpha\in(0,1)$,
%failure rate $\delta\in(0,1)$.
%\State
%$\forall l=1,\dots,m: 
%\hat{F}_l(\vec{x}) \defn \frac{1}{n}\sum_{i=1}^n \mathds{1}\{s_l (\vec{x}_i) \leq s_l(\vec{x})\}
%\text{~where~}
%\{\vec{x}_i\}_{i=1}^n = \mc{X}\train 
%$
%\State
%$\forall l=1,\dots,m: \hat{z}_l(\vec{x}) \defn \Phi^{-1}(\hat{F}_l(\vec{x})) 
%$
%\State
%$\scoreglrt(\vec{x}) \defn \sum_{l=1}^m \big(\frac{1}{2}\hat{z}_l^-(\vec{x}) -\hat{z}_l(\vec{x})\big)\hat{z}_l^-(\vec{x})$ for $\hat{z}_l^-(\vec{x})=\min\{\hat{z}_l(\vec{x}),-\epsilon\}$
%\State
%$c_v(\delta) \defn \frac{-\ln(-\ln(1-\delta))+2\ln\ln v+0.5\ln\ln\ln v-0.5\ln\pi}{\sqrt{2\ln\ln n}}$ for $v\defn |\mc{X}\val|$
%\State
%$\forall i=1,\dots,v: b_i=\min\{1,\frac{i}{v}+c_v(\delta)\frac{\sqrt{i(v+1)}}{v\sqrt{v}}\}$
%\State
%$\breve{\qt}(\vec{x}) = b_{\lceil (v+1)\hat{\qt}(\vec{x})\rceil}$
%for 
%$\displaystyle \hat{\qt}(\vec{x}) = \frac{1+|\{\vec{x}_i\in\mc{X}\val: \scoreglrt(\vec{x}_i)\leq \scoreglrt(\vec{x})\}|}{1+v}$ 
%\State
%\Return ``OOD'' if $\breve{\qt}(\vec{x})\leq \alpha$, else ``inlier.''
%\end{algorithmic}
%\end{algorithm}

\subsection{Using the GLRT for self-supervised OOD detection} 

%\Secref{self} reviewed the state-of-the-art self-supervised CSI and SupCSI OOD-detection techniques from \citep{Tack:NIPS:20}.
%CSI can be used when the inliers have no class labels, while SupCSI exploits inlier class labels to achieve better performance.

As discussed in \secref{self}, many self-supervised OOD detectors compute intermediate scores using pretext tasks, augmented views, and/or distributions shifts, and then combine them heuristically \citep{Golan:NIPS:18,Hendrycks:NIPS:19,Bergman:ICLR:20,Tack:NIPS:20,Georgescu:CVPR:21,Khalid:CVPRW:22}.
We instead propose to combine these scores using our GLRT technique.
Although our approach could in principle be used to combine any set of inlier scores, we focus on the scores generated by the contrastive CSI and SupCSI techniques from \citep{Tack:NIPS:20} due to their state-of-the-art performance.

As described in \secref{self}, CSI \citep{Tack:NIPS:20} uses three ``base'' score functions,
%\eqref{cosj}-\eqref{shiftj} 
$\scorecosj(\cdot)$, $\scorenormj(\cdot)$, and $\scoreshiftj(\cdot)$,
evaluating each on $|\mc{T}|=4$ shifts and combining the $12$ resulting scores using the heuristic in \eqref{csi}.
We propose to instead combine these $12$ scores using the GLRT approach from \secref{glrt}.
We will refer to this latter approach as ``glrt-CSI.''

In the case where the inliers have class labels, one can use SupCSI \citep{Tack:NIPS:20} to outperform CSI.
%As described in \secref{self}, 
SupCSI's score \eqref{supcsi} ensembles the logits $\vec{\ell}(\cdot)\in\Real^{K\times |\mc{T}|}$ of a joint class/shift classifier head across the $|\mc{T}|$ shifts before softmax thresholding.
We instead propose to compute a separate softmax-thresholding score for each shift $j$, i.e.,
\begin{align}
\scoresupcsij(\vec{x})
&\defn \max_{k=1,\dots,K} \big[\softmax(\vec{\ell}_{:,j}(R_j(\vec{x})))\big]_k,
\quad j=1...|\mc{T}|
\label{eq:supcsij} ,
\end{align}
and combine them using the GLRT approach from \secref{glrt}, yielding what we call ``glrt-SupCSI.''

Finally, we propose a method that GLRT-combines the CSI and SupCSI base scores, in addition to incorporating two additional base scores:
i) one based on supervised OC-SVM 
\begin{align}
%\scoresupocsvm(\vec{x}) 
%&= \textstyle 
%\max_{k=1,\dots,K} \big[
%\sum_{\vec{x}_i\in\mc{X}\traink} \alpha_{ki}\, \kappa(\vec{x}_i,\vec{x}) - \rho_k
%\big] \\
%\scoresupocsvm(\vec{x}) 
%&= \textstyle 
%\max_{k=1,\dots,K} \sum_j \big[
%\sum_{\vec{x}_i\in\mc{X}\traink} \alpha_{kji}\, \kappa(R_j(\vec{x}_i),R_j(\vec{x})) - \rho_{kj}
%\big] \\
\scoresupocsvmj(\vec{x}) 
&\defn 
\max_{k=1,\dots,K} \left[
\sum_{\vec{x}_i\in\mc{X}\traink} \alpha_{kji}\, \kappa\big(
f_{\vec{\theta}}(R_j(\vec{x}_i)), f_{\vec{\theta}}(R_j(\vec{x}))
\big) - \rho_{kj}
\right]
%~j=1...|\mc{T}|
\label{eq:supocsvm},
\end{align}
where $\kappa(\vec{g},\vec{g}') \defn e^{-\gamma\|\vec{g}-\vec{g}'\|^2}$ is the radial basis function kernel and 
$\{\alpha_{kji}\}$ and $\rho_{kj}$ are parameters trained on $R_j(\vec{x}_i)$ for the class-$k$ subset of $\vec{x}_i\in\mc{X}\train$,
and ii) one based on supervised Mahalanobis 
\begin{align}
%\scoresupmah(\vec{x}) 
%&= \textstyle -\min_{k=1,\dots,K} \frac{1}{|\mc{T}|}\sum_j (g_{\vec{\theta}}(R_j(\vec{x})) - \hvec{\mu}_{kj})\hvec{\Sigma}_{kj}^{-1} (g_{\vec{\theta}}(R_j(\vec{x})) - \hvec{\mu}_{kj}) \\
\scoresupmahj(\vec{x}) 
&\defn -\min_{k=1,\dots,K}\,\left[ (h_{\vec{\theta}}(R_j(\vec{x})) - \hvec{\mu}_{kj})\hvec{\Sigma}_{kj}^{-1} (h_{\vec{\theta}}(R_j(\vec{x})) - \hvec{\mu}_{kj})
+ \log |\hvec{\Sigma}_{kj}| \right]
\label{eq:supmah} ,
\end{align}
for $j=1,\dots,|\mc{T}|$,
where $h_{\vec{\theta}}(\cdot)$ is the penultimate layer of a ResNet18 trained (with label smoothing \citep{Szegedy:CVPR:16}) to jointly classify the shift $j\in\mc{T}$ and inlier class $k$.
We found that using this network to compute $\scoresupmahj$ worked slightly better than using the SupCSI network. 
In \eqref{supmah}, $\hvec{\mu}_{kj}$ and $\hvec{\Sigma}_{kj}$ are the sample mean and covariance of $\{h_{\vec{\theta}}(R_j(\vec{x}_i))\}$ for $\{\vec{x}_i\}$ in the class-$k$ subset of $\mc{X}\train$.
This latter approach, which combines 24 scores in total, is called ``glrt-SupCSI+.''
See \tabref{methods} for a summary.

\putTable{methods}
{Self-supervised OOD detectors and their base scores}
{\resizebox{\columnwidth}{!}{%
  \begin{tabular}{@{}lcccccccc@{}}
    \toprule
    & \multicolumn{6}{c}{Base Scores} & \multicolumn{2}{c}{Combining} \\
    \cmidrule(r){2-7}
    \cmidrule(r){8-9}
    Method & $\scorecosj$ & $\scorenormj$ & $\scoreshiftj$ & %$\scoreocsvmj$ & $\scoremahj$ & 
        $\scoresupcsij$ & $\scoresupocsvmj$ & $\scoresupmahj$ & Heuristic & GLRT\\
    \midrule
    CSI \citeo{Tack:NIPS:20} & \checkmark & \checkmark & \checkmark & & & & \checkmark & \\
    glrt-CSI & \checkmark & \checkmark & \checkmark & & & & & \checkmark \\
    %glrt-CSI+ & \checkmark & \checkmark & \checkmark & ? & ? & & & & \checkmark \\
    \midrule 
    SupCSI \citeo{Tack:NIPS:20} & & & & \checkmark & & & \checkmark & \\
    glrt-SupCSI & & & & \checkmark & & & & \checkmark \\
    glrt-SupCSI+ & \checkmark & \checkmark & \checkmark & \checkmark & \checkmark & \checkmark & & \checkmark \\
    \bottomrule
 \end{tabular} 
}}

%*******************************************************************************
\section{Experimental results} \label{sec:experiments}

We now present experimental results comparing our proposed GLRT-based methods to CSI and SupCSI from \citep{Tack:NIPS:20} using image data.
For the case where the inliers have class labels, we compare to two additional baselines: the softmax-thresholding approach from \citep{Hendrycks:ICLR:17} and the Mahalanobis-based approach from \citep{Lee:NIPS:18}, which both use a supervised classification DNN trained using cross-entropy loss.
In addition, we compare our GLRT-based score-combining method to the traditional score-combining methods of Fisher, Bonferroni, Simes/BH, Stouffer, and ALR \citep{Walther:IMSC:13} using the SupCSI+ configuration.

In the subsections below, we focus on two OOD-detection tasks.
In the standard ``dataset-vs-dataset'' task, the inliers are defined by one dataset (e.g., CIFAR-10) and the novelties are defined by a different dataset (e.g., SVHN).  
For this task, we evaluate 8 different inlier/novelty combinations.
In the ``leave-one-class-out'' task, the inliers are defined by 9 classes of CIFAR-10 and the novelties are defined by the remaining 10th class.
%For this task, we evaluate all 10 possible configurations.
This latter task is considered to be much more practically relevant \citep{Ahmed:AAAI:20} 
than the dataset-vs-dataset task or the ``leave-one-class-in'' task, where the inliers are defined by 1 class and the novelties by the remaining 9 classes. 
To evaluate performance, we will consider both detection rate (DR) for a fixed false-alarm rate (FAR), as well as the area under the receiver operating curve (AUROC), i.e., area under the DR-versus-FAR curve.
For the DR-versus-FAR experiments, we used the minimal detection threshold $\tau$ under which $\alpha\test(\tau)\leq\alpha$, where $\alpha$ is the target FAR and $\alpha\test(\tau)$ is the FAR measured on the test set.

\subsection{Implementation details}

First we describe the training of the four DNNs used respectively for CSI, SupCSI, joint class/shift classification (for $\scoresupmahj$), and supervised classification.
All models use the ResNet18 architecture \citep{He:CVPR:16} for feature encoding.
CSI and SupCSI are trained using the authors' code under default parameter settings.
The joint classifier network is trained using cross-entropy loss with label smoothing \citep{Szegedy:CVPR:16} and joint class-shift labels.
Finally, the supervised classification network is trained using cross-entropy loss with inlier class labels and without the use of distributional shifting transforms.
All networks are trained for 100 epochs using the LARS \citep{You:17} optimizer.
The learning rate is scheduled using cosine annealing \citep{Smith:SPIE:19} with initial value 0.1, max value 1.0, one epoch of warmup, and a final value of 10$^{-6}$.
The batch size is 128.
%For all experiments, the batch size was set to $128 / |\mc{T}|$, which makes the batch augmented with the shifting transforms have a size of $128$. 
%The training time scales with the number of shifts $|\mc{T}|$, in that training a model with $n_r=4$ takes four times as long as training a model with $n_r=1$.

\begin{figure}[t]
\includegraphics[width=2.5in,trim=7 8 8 7,clip]{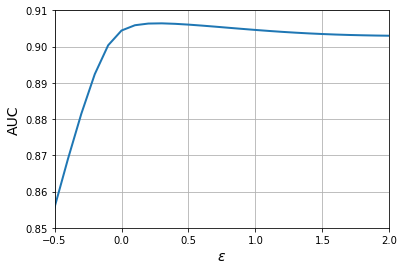}
\caption{AUROC of glrt-SupCSI+ vs.\ GLRT parameter $\epsilon$ for CIFAR-10 inliers and CIFAR-100 novelties.}
\label{fig:epsilon}
\end{figure}

\subsection{Choice of $\epsilon$} \label{sec:epsilon}
%\textbf{Choice of $\epsilon$.}
We first consider the choice of $\epsilon$ in the NM problem \eqref{NMP} on which our GLRT is based. 
\Figref{epsilon} shows AUROC versus $\epsilon$ for the specific task of distinguishing CIFAR-100 novelties from CIFAR-10 inliers.
There it can be seen that values of $\epsilon\in (0,1)$ slightly outperform the conventional choice \citep{Wei:AS:19} of $\epsilon=0$ for this combination of task and metric.
Although better performance could be achieved by individually optimizing $\epsilon$ for each combination of the 18 tasks and two metrics (DR and AUROC) that we consider, doing would require access to OOD data that is unavailable in practice.
Thus, we use $\epsilon=0.25$ in all cases.

\subsection{Dataset-vs-dataset experiments} \label{sec:dvd}
%\textbf{Dataset-vs-dataset experiments.}
In this section, we consider experiments where i) the inliers are CIFAR-10 \citep{Krizhevsky:CIFAR:09} and the novelties are either SVHN \citep{Netzer:NIPS:11}, LSUN \citep{Yu:LSUN:15}, ImageNet \citep{Deng:CVPR:09}, or CIFAR-100 \citep{Krizhevsky:CIFAR:09}, and ii) where the inliers are SVHN and the novelties are either LSUN, ImageNet, CIFAR-10, or CIFAR-100.
The OOD detectors are trained using the training fold of the inlier dataset and tested using the testing folds of the inlier and novelty datasets.

Tables \ref{tab:dvd_auroc_cifar10}-\ref{tab:dvd_auroc_svhn} show AUROC for several techniques, while Tables~\ref{tab:dvd_pdpfa_cifar10}-\ref{tab:dvd_pdpfa_svhn} show DR for a fixed FAR of 5\%. 
In all of these tables, the techniques above the horizontal line (i.e., CSI and glrt-CSI) have no access to inlier class labels, while the techniques below the horizontal line do.
The tables show the contrastive CSI and SupCSI techniques outperforming the softmax-thresholding and Mahalanobis baselines noticeably in AUROC, and drastically in DR, consistent with the existing literature (e.g., \citep{Tack:NIPS:20}).
The tables also show the proposed glrt-CSI and glrt-SupCSI outperforming their CSI and SupCSI counterparts on average in both AUROC and DR. 
This demonstrates the advantage of GLRT score-combining over the heuristic combining in \eqref{csi}-\eqref{supcsi}.
Finally, the tables show that, compared to glrt-SupCSI, the proposed glrt-SupCSI+ does significantly better in DR for CIFAR-10 inliers, slightly better in AUROC in both experiments, and slightly worse in DR for SVHN inliers.
This shows that incorporating additional scores \emph{can} be very advantageous, but is not guaranteed to be.
We will discuss this phenomenon further in the context of the leave-one-out experiments.

\putTable{dvd_auroc_cifar10}{Dataset-vs-dataset AUROC for CIFAR-10 inliers}{
    \scalebox{0.9}{%
        \begin{tabular}{l c c c c c}
            \toprule
            & \multicolumn{4}{c}{Novelty Dataset} & \\
            \cmidrule(r){2-5}
            Method & SVHN & LSUN & ImageNet & CIFAR-100 & Avg.\\
            \midrule
            CSI \citeo{Tack:NIPS:20} & \bf 0.9950 & 0.9750 & 0.9783 & 0.8628 & 0.9528 \\
            %\multirow{2}{*}{No} & CSI \citep{Tack:NIPS:20} & 0.9979 & 0.9713 & 0.9756 & 0.8848 & 0.9574 \\ % w/ens
            glrt-CSI & 0.9947 & \bf 0.9768 & \bf 0.9784 & \bf 0.8709 & \bf 0.9552 \\
            \midrule
            %\multirow{4}{*}{Yes} & Mah \citep{Lee:NIPS:18} & 0.9041 & 0.7175 & 0.6209 & 0.7077 & 0.7375 \\
            %\multirow{4}{*}{Yes} & Mah \citep{Lee:NIPS:18} & 0.5501 & 0.5979 & 0.4910 & 0.6832 & 0.5806 \\
            Softmax \citeo{Hendrycks:ICLR:17} & 0.9637 & 0.9015 & 0.9006 & 0.8643 & 0.9075 \\
            Mah \citeo{Lee:NIPS:18} & 0.6063 & 0.5767 & 0.4688 & 0.6923 & 0.5860 \\
            SupCSI \citeo{Tack:NIPS:20} & 0.9844 & 0.9787 & 0.9788 & 0.9191 & 0.9652 \\
            %NCIS-CSI & 0.9892 & 0.9817 & 0.9838 & 0.8479 & ? \\
            %NCIS-CSI & 0.9974 & 0.9712 & 0.9736 & 0.8794 & 0.9554 \\ % w/ens
            %SEND-CSI & 0.9950 & 0.9768 & 0.9787 & 0.8584 & ? \\
            %SEND-CSI & 0.9950 & 0.9768 & 0.9787 & 0.8638 & 0.9536 \\
            %SEND-CSI & 0.9982 & 0.9694 & 0.9726 & 0.8820 & 0.9556 \\ % w/ens
            glrt-SupCSI & 0.9863 & 0.9808 & 0.9812 & \bf 0.9214 & 0.9674 \\
            glrt-SupCSI+ & \bf 0.9968 & \bf 0.9836 & \bf 0.9840 & 0.9064 & \bf 0.9677 \\
                      %& Full-SEND & 0.9967 & 0.9834 & 0.9838 & 0.9071 & 0.9678 \\ % no ocsvm, joint-mah
            \bottomrule
        \end{tabular}%
    }
}
\putTable{dvd_auroc_svhn}{Dataset-vs-dataset AUROC for SVHN inliers}{
    \scalebox{0.9}{%
        \begin{tabular}{l c c c c c}
            \toprule
            & \multicolumn{4}{c}{Novelty Dataset} & \\
            \cmidrule(r){2-5}
            Method & LSUN & ImageNet & CIFAR-10 & CIFAR-100 & Avg.\\
            \midrule
            CSI \citeo{Tack:NIPS:20} & 0.9571 & 0.9631 & 0.9638 & 0.9556 & 0.9599 \\
            %CSI \citep{Tack:NIPS:20} & 0.8919 & 0.8989 & 0.9638 & 0.9597 & 0.9286 \\ % w/ens
            glrt-CSI & \bf 0.9776 & \bf 0.9808 & \bf 0.9677 & \bf 0.9577 & \bf 0.9710 \\
            \midrule
            %\multirow{4}{*}{Yes} & Mah \citep{Lee:NIPS:18} & 0.9069 & 0.9122 & 0.7809 & 0.7692 & 0.8423 \\
            %\multirow{4}{*}{Yes} & Mah \citep{Lee:NIPS:18} & 0.9478 & 0.9562 & 0.9371 & 0.9357 & 0.9442 \\
            Softmax \citeo{Hendrycks:ICLR:17} & 0.9310 & 0.9364 & 0.9346 & 0.9274 & 0.9324 \\
            Mah \citeo{Lee:NIPS:18} & 0.9532 & 0.9616 & 0.9427 & 0.9397 & 0.9493 \\
            %NCIS-CSI & 0.9441 & 0.9493 & 0.9748 & 0.9643 & ? \\
            %NCIS-CSI & 0.9784 & 0.9767 & 0.9816 & 0.9746 & 0.9778 \\ % w/ens
            %SEND-CSI & 0.9888 & 0.9890 & 0.9801 & 0.9726 & ? \\
            %SEND-CSI & 0.9841 & 0.9835 & 0.9839 & 0.9782 & 0.9824 \\ % w/ens
            SupCSI \citeo{Tack:NIPS:20} & 0.9882 & 0.9904 & 0.9909 & 0.9860 & 0.9889 \\
            glrt-SupCSI & 0.9917 & 0.9929 & \bf 0.9930 & \bf 0.9887 & 0.9916 \\
            glrt-SupCSI+ & \bf 0.9945 & \bf 0.9952 & 0.9927 & \bf 0.9887 & \bf 0.9928 \\ % no ens
                      %& Full-SEND & 0.9913 & 0.9927 & 0.9901 & 0.9853 & 0.9899 \\ % no ocsvm, joint-mah
            %Full-SEND & \bf 0.9928 & \bf 0.9932 & \bf 0.9936 & \bf 0.9905 & \bf 0.9925 \\ % w/ens
            \bottomrule
        \end{tabular}%
    }%
}

\putTable{dvd_pdpfa_cifar10}{Dataset-vs-dataset detection rate (\%) at a false-alarm rate of 5\% for CIFAR-10 inliers}{
    \scalebox{0.9}{%
        \begin{tabular}{l c c c c c}
            \toprule
            & \multicolumn{4}{c}{Novelty Dataset} & \\
            \cmidrule(r){2-5}
            Method & SVHN & LSUN & ImageNet & CIFAR-100 & Avg.\\
            \midrule
            CSI \citeo{Tack:NIPS:20} & \bf 98.6 & 87.4 & 90.1 & \bf 45.3 & 80.4 \\
            %\multirow{2}{*}{No} & CSI \citep{Tack:NIPS:20} & 95.9 & 47.2 & 52.0 & 21.6 & 54.2 \\ % w/ens
            glrt-CSI & 98.2 & \bf 90.9 & \bf 92.0 & 45.2 & \bf 81.6 \\
            \midrule
            %\multirow{4}{*}{Yes} & Mah \citep{Lee:NIPS:18} & 10.2 & 5.2 & 4.6 & 20.7 & 10.2 \\
            Softmax \citeo{Hendrycks:ICLR:17} & 74.9 & 40.6 & 38.7 & 32.6 & 46.7 \\
            Maha \citeo{Lee:NIPS:18} & 11.7 & 5.1 & 4.3 & 21.2 & 10.6 \\
            SupCSI \citeo{Tack:NIPS:20} & 91.9 & 88.5 & 89.5 & \bf 59.8 & 82.4 \\
            %NCIS-CSI & ? & ? & ? & ? & ? \\
            %NCIS-CSI & 93.8 & 45.0 & 39.2 & 21.6 & 49.9 \\ % w/ens
            %SEND-CSI & 86.8 & 39.0 & 38.5 & 17.6 & 45.5 \\
            %SEND-CSI & 95.4 & 44.9 & 40.8 & 21.9 & 50.8 \\ % w/ens
             %& SEND-SupCSI & 92.7 & 89.6 & 90.3 & 57.6 & ? \\ % eps=0.
            glrt-SupCSI & 92.7 & 89.6 & 90.3 & 57.6 & 82.6 \\ % eps=0.25
            glrt-SupCSI+ & \bf 99.3 & \bf 95.4 & \bf 95.7 & 54.3 & \bf 86.2 \\
             %& Full-SEND & 99.2 & 95.3 & 95.5 & 54.0 & 86.0 \\ % no ocsvm, joint-mah
                        %& Full-SEND & \bf 96.6 & 55.6 & 52.24 & \bf 24.7 & \bf 57.3 \\ %w/ens
            \bottomrule
        \end{tabular}%
    }
}
\putTable{dvd_pdpfa_svhn}{Dataset-vs-dataset detection rate (\%) at a false-alarm rate of 5\% for SVHN inliers}{
    \scalebox{0.9}{%
        \begin{tabular}{l c c c c c}
            \toprule
            & \multicolumn{4}{c}{Novelty Dataset} & \\
            \cmidrule(r){2-5}
            Method & LSUN & ImageNet & CIFAR-10 & CIFAR-100 & Avg.\\
            \midrule
            CSI \citeo{Tack:NIPS:20} & 70.0 & 76.5 & 78.8 & 74.6 & 75.0 \\
            %\multirow{2}{*}{No} & CSI \citep{Tack:NIPS:20} & 95.9 & 47.2 & 52.0 & 21.6 & 54.2 \\ % w/ens
            glrt-CSI & \bf 89.5 & \bf 91.7 & \bf 81.6 & \bf 75.9 & \bf 84.7 \\
            \midrule
            %\multirow{4}{*}{Yes} & Mah \citep{Lee:NIPS:18} & 66.4 & 72.3 & 58.7 & 58.8 & 64.1 \\
            Softmax \citeo{Hendrycks:ICLR:17} & 58.6 & 62.5 & 62.8 & 61.7 & 61.4 \\
            Maha \citeo{Lee:NIPS:18} & 70.0 & 76.3 & 61.9 & 62.0 & 67.6 \\
            SupCSI \citeo{Tack:NIPS:20} & 97.1 & 98.2 & 97.9 & 94.4 & 96.9 \\
             %& SEND-SupCSI & 98.4 & 99.0 & 98.7 & 95.8 & ? \\ % eps=0.0
            glrt-SupCSI & 98.4 & 99.0 & \bf 98.7 & \bf 95.8 & \bf 98.0 \\ % eps=0.25
            glrt-SupCSI+ & \bf 99.3 & \bf 99.2 & 97.7 & 94.9 & 97.8 \\
             %& Full-SEND & 98.4 & 98.7 & 96.1 & 92.5 & 96.4 \\ % no ocsvm, joint-mah
            \bottomrule
        \end{tabular}%
    }%
}

Tables~\ref{tab:dvd_auroc_scores_cifar10}-\ref{tab:dvd_pdpfa_scores_svhn} compare the proposed GLRT to several existing score-combining rules using the AUROC and DR metrics, respectively.
In terms of AUROC, Tables~\ref{tab:dvd_auroc_scores_cifar10}-\ref{tab:dvd_auroc_scores_svhn} show the GLRT outperforming all other methods on average, with Fisher and Stouffer close behind and the other methods performing much worse.
In terms of DR, \tabref{dvd_pdpfa_scores_cifar10} shows that, with CIFAR-10 inliers, Stouffer's method wins and the GLRT and Fisher tie for a close 2nd place, while \tabref{dvd_pdpfa_scores_svhn} shows that, with SVHN inliers, the GLRT and Fisher tie for 1st place while Stouffer is close behind.
All together, for the dataset-vs-dataset experiments, the GLRT shows a slight advantage over Fisher and Stouffer and a major advantage over Bonferroni, Simes/BH, and ALR.

\putTable{dvd_auroc_scores_cifar10}{Dataset-vs-dataset AUROC of SupCSI+ with different score-combining techniques and CIFAR-10 inliers}{%
    %\scalebox{0.7}{
    %\begin{tabular}{l c c c c c c c c c}
        %Inlier Dataset & \multicolumn{4}{c|}{CIFAR-10} & \multicolumn{4}{c|}{SVHN} & \\
        %Novelty Dataset & SVHN & LSUN & ImageNet & CIFAR-100 & LSUN & ImageNet & CIFAR-10 & CIFAR-100 & Avg. \\
        %\hline
        %%Full-SEND & 0.9195, 0.9928, 0.9986 & 0.5655, 0.9536, 0.9821 & 0.5453, 0.9567, 0.984 & 0.2166, 0.5426, 0.7077 & 0.8704, 0.9934, 0.9989 & 0.9049, 0.9919, 0.9980 & 0.8147, 0.9771, 0.9947 & 0.7424, 0.9491, 0.9869 & ? \\
        %Full-SEND & 0.9968 & 0.9836 & 0.9840 & 0.9064 & 0.9945 & 0.9952 & 0.9927 & 0.9887 & 0.9802 \\ 
        %Fisher & 0.9973 & 0.9839 & 0.9838 & 0.9031 & 0.9942 & 0.9950 & 0.9925 & 0.9887 & 0.9798 \\ 
        %Bonferroni & 0.9833 & 0.9719 & 0.9759 & 0.8906 & 0.9928 & 0.9929 & 0.9865 & 0.9811 & 0.9719 \\ 
        %Simes & 0.9837 & 0.9730 & 0.9767 & 0.8920 & 0.9929 & 0.9931 & 0.9873 & 0.9823 & 0.9726 \\ 
        %ALR & 0.9959 & 0.9591 & 0.9541 & 0.8282 & 0.8331 & 0.8462 & 0.9047 & 0.9146 & 0.9045 \\ 
        %Stouffer & 0.9976 & 0.9842 & 0.9830 & 0.9033 & 0.9929 & 0.9939 & 0.9916 & 0.9880 & 0.9793 \\ 
    %\end{tabular}
%
    \scalebox{0.9}{%
        \begin{tabular}{l c c c c c}
            %\multicolumn{6}{c}{(a) Inlier dataset: CIFAR-10}\\
            \toprule
            & \multicolumn{4}{c}{Novelty Dataset} & \\
            \cmidrule(r){2-5}
            Method & SVHN & LSUN & ImageNet & CIFAR-100 & Avg.\\
            \midrule
            GLRT & 0.9968 & 0.9836 & \bf 0.9840 & \bf 0.9064 & \bf 0.9677 \\ 
            Fisher & \bf 0.9973 & 0.9839 & 0.9838 & 0.9031 & 0.9670 \\
            Bonferroni & 0.9833 & 0.9719 & 0.9759 & 0.8906 & 0.9554 \\
            Simes/BH & 0.9837 & 0.9730 & 0.9767 & 0.8920 & 0.9564 \\
            ALR & 0.9959 & 0.9591 & 0.9541 & 0.8282 & 0.9343 \\
            Stouffer & 0.9976 & \bf 0.9842 & 0.9830 & 0.9033 & 0.9670 \\ 
            \bottomrule
        \end{tabular}%
    }
}
\putTable{dvd_auroc_scores_svhn}{Dataset-vs-dataset AUROC of SupCSI+ with different score-combining techniques and SVHN inliers}{%
    \scalebox{0.9}{%
        \begin{tabular}{l c c c c c}
            %\multicolumn{6}{c}{(b) Inlier dataset: SVHN}\\
            \toprule
            & \multicolumn{4}{c}{Novelty Dataset} & \\
            \cmidrule(r){2-5}
            Method & LSUN & ImageNet & CIFAR-10 & CIFAR-100 & Avg.\\
            \midrule
            GLRT & \bf 0.9945 & \bf 0.9952 & \bf 0.9927 & \bf 0.9887 & \bf 0.9928 \\ 
            Fisher & 0.9942 & 0.9950 & 0.9925 & \bf 0.9887 & 0.9926 \\ 
            Bonferroni & 0.9928 & 0.9929 & 0.9865 & 0.9811 & 0.9883 \\ 
            Simes/BH & 0.9929 & 0.9931 & 0.9873 & 0.9823 & 0.9889 \\ 
            ALR & 0.8331 & 0.8462 & 0.9047 & 0.9146 & 0.8747 \\ 
            Stouffer & 0.9929 & 0.9939 & 0.9916 & 0.9880 & 0.9916 \\ 
            \bottomrule
        \end{tabular}%
    }
}

\putTable{dvd_pdpfa_scores_cifar10}{Dataset-vs-dataset detection rate (\%) at a false alarm rate of 5\% for SupCSI+ with different score-combining techniques and CIFAR-10 inliers}{%
    %\scalebox{0.7}{
    %\begin{tabular}{l c c c c c c c c c}
        %Inlier Dataset & \multicolumn{4}{c|}{CIFAR-10} & \multicolumn{4}{c|}{SVHN} & \\
        %Novelty Dataset & SVHN & LSUN & ImageNet & CIFAR-100 & LSUN & ImageNet & CIFAR-10 & CIFAR-100 & Avg. \\
        %\hline
        %Full-SEND & 0.9195, 0.9928, 0.9986 & 0.5655, 0.9536, 0.9821 & 0.5453, 0.9567, 0.984 & 0.2166, 0.5426, 0.7077 & 0.8704, 0.9934, 0.9989 & 0.9049, 0.9919, 0.9980 & 0.8147, 0.9771, 0.9947 & 0.7424, 0.9491, 0.9869 & ? \\
        %Fisher & 0.9315, 0.9932, 0.9988 & 0.45693, 0.9553, 0.9823 & 0.5239, 0.9572, 0.9836 & 0.2231, 0.5435, 0.703 & 0.8745, 0.9942, 0.9992 & 0.9081, 0.9921, 0.9981 & 0.8172, 0.9766, 0.9944 & 0.7462, 0.9490, 0.9872 & ? \\
        %Bonferroni & 0.0, 0.9603, 0.9922 & 0.0,94.90.863, 0.9581 & 0.0, 0.8948, 0.9655 & 0.0, 0.4512, 0.6503 & 0.8546, 0.9837, 0.9962 & 0.8776, 0.9829, 0.9953 & 0.7285, 0.9346, 0.9801 & 0.6565, 0.8934, 0.9624 & ? \\
        %Simes & 0.0, 0.9659, 0.9937 & 0.0, 0.8758, 0.9637 & 0.0, 0.9028, 0.9708 & 0.0, 0.4582, 0.6587 & 0.8508, 0.9843, 0.9970 & 0.8761, 0.9842, 0.9957 & 0.7333, 0.9424, 0.9854 & 0.6594, 0.8993, 0.9684 & ? \\
    \scalebox{0.9}{%
        \begin{tabular}{l c c c c c}
            %\multicolumn{6}{c}{(a) Inlier dataset: CIFAR-10} \\
            \toprule
            & \multicolumn{4}{c}{Novelty Dataset} & \\
            \cmidrule(r){2-5}
            Method & SVHN & LSUN & ImageNet & CIFAR-100 & Avg.\\
            \midrule
            GLRT& 99.3 & 95.4 & \bf 95.7 & 54.3 & 86.2 \\
            Fisher & 99.3 & \bf 95.5 & \bf 95.7 & 54.4 & 86.2 \\
            Bonferroni & 96.0 & 86.3 & 89.5 & 45.1 & 79.2 \\
            Simes/BH & 96.6 & 87.6 & 90.3 & 45.8 & 80.1 \\
            ALR & 97.9 & 86.3 & 79.6 & 44.6 & 77.1 \\
            Stouffer & \bf 99.4 & \bf 95.5 & \bf 95.7 & \bf 54.6 & \bf 86.3 \\ 
            \bottomrule
        \end{tabular}%
    }
}

\putTable{dvd_pdpfa_scores_svhn}{Dataset-vs-dataset detection rate (\%) at a false alarm rate of 5\% for SupCSI+ with different score-combining techniques and SVHN inliers}{%
    \scalebox{0.9}{%
        \begin{tabular}{l c c c c c}
            %\multicolumn{6}{c}{(b) Inlier dataset: SVHN} \\
            \toprule
            & \multicolumn{4}{c}{Novelty Dataset} & \\
            \cmidrule(r){2-5}
            Method & LSUN & ImageNet & CIFAR-10 & CIFAR-100 & Avg.\\
            \midrule
            GLRT & 99.3 & \bf 99.2 & \bf 97.7 & \bf 94.9 & \bf 97.8 \\
            Fisher &  \bf 99.4 & \bf 99.2 & \bf 97.7 & \bf 94.9 & \bf 97.8 \\
            Bonferroni & 98.4 & 98.3 & 93.5 & 89.3 & 94.9 \\
            Simes/BH &  98.4 & 98.4 & 94.2 & 89.9 & 95.2 \\
            ALR & 38.6 & 44.6 & 54.5 & 59.0 & 49.2 \\
            Stouffer & \bf 99.4 & 99.1 & 97.3 & 94.5 & 97.6 \\ 
            \bottomrule
        \end{tabular}%
    }
}

\subsection{Leave-one-class-out experiments} \label{sec:loo}
%\textbf{Leave-one-class-out experiments.}
We now consider experiments where the inliers are 9 classes of CIFAR-10 and the novelties are the remaining class.
All 10 configurations are considered under both the AUROC and DR metrics.
The standard CIFAR-10 train and test folds are used for training and testing, respectively.

\tabref{loo_auroc} shows AUROC for several techniques, while \tabref{loo_pdpfa} shows DR for a fixed FAR of 25\%.
(The FAR was increased to yield meaningful DR values.)
The tables show SupCSI outperforming the softmax-thresholding and Mahalanobis baselines noticeably in both AUROC and DR. 
They also show the proposed glrt-CSI and glrt-SupCSI outperforming their CSI and SupCSI counterparts on average in both AUROC and DR, demonstrating the advantage of GLRT combining. 
However, the tables show glrt-SupCSI+ underperforming glrt-SupCSI, and even SupCSI, in both AUROC and DR. 
This is explained by \tabref{loo_base_scores}, which shows that, among the base scores, $\scoresupcsij$ has by far the highest AUROC in these leave-one-out experiments. 
Thus, combining $\scoresupcsij$ with other base scores (as in glrt-SupCSI+) leads to lower performance than with $\scoresupcsij$ alone (as in SupCSI and glrt-SupCSI).

\putTable{loo_auroc}{Leave-one-class-out AUROC}{
    \scalebox{0.73}{%
        \begin{tabular}{l c c c c c c c c c c c}
            \toprule
            &\multicolumn{10}{c}{CIFAR-10 Novelty Class} & \\
            \cmidrule(r){2-11}
            Method & Plane & Car & Bird & Cat & Deer & Dog & Frog & Horse & Ship & Truck & Avg.\\
            \midrule
        %Mah \citep{Lee:NIPS:18} & 0.9041 & 0.7175 & 0.6209 & 0.7077 & 0.9069 & 0.9122 & 0.7809 & 0.7692 & 0.7899 \\
        %OC-SVM \citep{Andrews:ICML:16} & 0.9813 & 0.7581 & 0.6998 & 0.6891 & 0.9929 & 0.9921 & 0.9862 & 0.9774 & 0.8846 \\
        %Norm & 0.7481 & 0.389 & 0.8308 & 0.8593 & 0.7423 & 0.7641 & 0.8266 & 0.6436 & 0.6873 & 0.5684 & ?\\
        %Cos & 0.7705 & 0.4869 & 0.8479 & 0.8784 & 0.7341 & 0.7992 & 0.8267 & 0.7053 & 0.6366 & 0.5645 & ?\\
        %Shift & 0.8259 & 0.4405 & 0.8835 & 0.8538 & 0.7302 & 0.7625 & 0.9245 & 0.5406 & 0.7007 & 0.5774 & ?\\
        %SupCSI-base & 0.8575 & 0.5868 & 0.9247 & 0.8902 & 0.9024 & 0.8607 & 0.9538 & 0.8198 & 0.7879 & 0.6564 & ?\\
        %OC-SVM \citep{Andrews:ICML:16} & 0.7216 & 0.5936 & 0.6418 & 0.6116 & 0.5451 & 0.5315 & 0.6629 & 0.5630 & 0.4702 & 0.6743 & 0.6016\\
         CSI \citeo{Tack:NIPS:20} & \bf 0.8230 & 0.4279 & \bf 0.8989 & 0.8889 & \bf 0.7582 & 0.7950 & \bf 0.9298 & 0.5893 & \bf 0.6983 & \bf 0.5876 & 0.7397 \\ % no ens
         %\multirow{2}{*}{No} & CSI \citep{Tack:NIPS:20} & \bf 0.8442 & 0.4448 & 0.9073 & \bf 0.9038 & 0.7512 & \bf 0.7894 & \bf 0.9439 & 0.6013 & \bf 0.7280 & \bf 0.6309 & 0.7545 \\ % w/ens
                            %& SEND-CSI & 0.8158 & 0.4306 & 0.8957 & 0.8929 & 0.7526 & 0.7984 & 0.9177 & 0.6343 & 0.6702 & 0.5752 & 0.7383 \\ % ONLY CSI Scores, w/out ens, eps=0.0
         glrt-CSI & 0.8185 & \bf 0.4385 & 0.8971 & \bf 0.8941 & 0.7577 & \bf 0.8008 & 0.9181 & \bf 0.6362 & 0.6765 & 0.5835 & \bf 0.7421 \\ % ONLY CSI Scores, w/out ens, eps=0.25
                            %& SEND-CSI & 0.8194 & 0.4176 & 0.8965 & 0.8850 & 0.7526 & 0.7846 & 0.9271 & 0.5933 & 0.6877 & 0.5802 & 0.7344 \\ % con and shift Scores, w/out ens, eps=0.25
                            %& SEND-CSI & 0.8403 & 0.4407 & 0.9074 & 0.9012 & 0.7495 & 0.7867 & 0.9353 & 0.6322 & 0.7015 & 0.6130 & 0.7508 \\ % ONLY CSI Scores, w/ ens, eps=0.0
                            %& SEND-CSI & 0.8430 & \bf 0.4537 & \bf 0.9087 & 0.9019 & \bf 0.7553 & 0.7885 & 0.9355 & \bf 0.6361 & 0.7061 & 0.6212 & \bf 0.7550 \\ % ONLY CSI Scores, w/ ens, eps=0.25
        \midrule
        %Joint-Mah & 0.9516 & 0.8023 & 0.7645 & 0.7907 & 0.9833 & 0.9871 & 0.9846 & 0.9801 \\
        %END-4 & 0.9892 & 0.9817 & 0.9838 & 0.8479 & 0.9441 & 0.9493 & 0.9748 & 0.9643 \\
        %SEND-16 & 0.9950 & 0.9768 & 0.9787 & 0.8584 & 0.9888 & 0.9890 & 0.9801 & 0.9726 \\
        %SEND-24 ($r_{max}=0.0$)& 0.8656 & 0.6244 & 0.9218 & 0.9067 & 0.8407 & 0.8299 & 0.9414 & 0.8091 & 0.7251 & 0.7193 & ? \\ % z_min = 0.0
        %NCIS-CSI & 0.8121 & 0.4606 & 0.8666 & 0.8695 & 0.7203 & 0.7698 & 0.8998 & 0.6581 & 0.6336 & 0.5984 & 0.7289 \\ %w/out ens
        %\multirow{4}{*}{Yes} & Mah \citep{Lee:NIPS:18} & 0.7186 & \bf 0.7710 & 0.6900 & 0.7123 & 0.5336 & 0.6180 & 0.4795 & 0.7714 & 0.6692 & \bf 0.7950 & 0.6759 \\
        Softmax \citeo{Hendrycks:ICLR:17} & 0.8304 & 0.6034 & 0.8528 & 0.8503 & 0.8621 & 0.8301 & 0.8669 & 0.8167 & 0.7727 & \bf 0.7199 & 0.8005 \\
        Maha \citeo{Lee:NIPS:18} & 0.7321 & \bf 0.6720 & 0.6943 & 0.7746 & 0.5714 & 0.7078 & 0.4909 & 0.7661 & 0.6694 & 0.6981 & 0.6777 \\
        %Mah \citep{Lee:NIPS:18} & 0.7304 & 0.7501 & 0.6662 & 0.7332 & 0.5187 & 0.5826 & 0.4080 & 0.7499 & 0.7137 & 0.7519 & 0.6605 \\
        SupCSI \citeo{Tack:NIPS:20} & 0.8827 & 0.5851 & 0.9437 & 0.9026 & 0.9231 & \bf 0.8793 & 0.9690 & \bf 0.8514 & 0.7913 & 0.6406 & 0.8369 \\
        %NCIS-CSI & 0.8361 & 0.4757 & 0.8922 & 0.8877 & 0.7256 & 0.7651 & 0.9287 & 0.6551 & 0.6733 & 0.6441 & 0.7484 \\ %w/ens
        %SEND-CSI & ? & ? & ? & ? & ? & ? & ? & ? & ? & ? & ? \\ % w/
        %SEND-CSI & ? & ? & ? & ? & ? & ? & ? & ? & ? & ? & ? \\ 
        %SEND-CSI & 0.8230 & 0.4656 & 0.8870 & 0.8861 & 0.7379 & 0.7904 & 0.9147 & 0.6731 & 0.6341 & 0.6143 & 0.7426 \\ %w/out ens
        %SEND-CSI & 0.8446 & 0.4789 & 0.8998 & 0.8945 & 0.7376 & 0.7767 & 0.9330 & 0.6726 & 0.6654 & 0.6443 & 0.7547 \\ % w/ ens
        glrt-SupCSI & \bf 0.8866 & 0.5971 & \bf 0.9448 & 0.9100 & \bf 0.9299 & 0.8729 & \bf 0.9704 & 0.8474 & \bf 0.8186 & 0.6745 & \bf 0.8452\\ % r_max = 0
        %SEND-SupCSI & 0.8859 & 0.5941 & \bf 0.9455 & 0.9107 & \bf 0.9296 & 0.8750 & \bf 0.9705 & 0.8474 & \bf 0.8141 & 0.6650 & \bf 0.8438 \\ % r_max = -1/4
        glrt-SupCSI+ & 0.8768 & 0.6099 & 0.9270 & \bf 0.9121 & 0.8511 & 0.8427 & 0.9426 & 0.8151 & 0.7497 & 0.7121 & 0.8239 \\ % w/out ens
                     %& Full-SEND & 0.8727 & 0.6073 & 0.9302 & 0.9145 & 0.8562 & 0.8451 & 0.9436 & 0.8060 & 0.7627 & 0.7204 & 0.8259 \\ % w/out ens, no ocsvm
                     %& Full-SEND & 0.8802 & 0.6096 & 0.9246 & 0.9055 & 0.8523 & 0.8400 & 0.9417 & 0.7990 & 0.7458 & 0.7122 & 0.8211 \\ % w/out ens, CSI-mah dist
                     %& Full-SEND & 0.8779 & 0.6085 & 0.9283 & 0.9143 & 0.8582 & 0.8428 & 0.9433 & 0.7899 & 0.7606 & 0.7049 & 0.8229 \\ % w/out ens, CSI-mah dist, no ocsvm
                     %& Full-SEND & 0.8686 & 0.5271 & 0.9330 & 0.9162 & 0.8530 & 0.8496 & 0.9558 & 0.7661 & 0.7606 & 0.6371 & 0.8067 \\ % w/out ens, CSI-mah dist, no ocsvm, no mah
                   %& Full-SEND & 0.8860 & 0.6128 & 0.9317 & \bf 0.9158 & 0.8514 & 0.8355 & 0.9514 & 0.8136 & 0.7592 & 0.7238 & 0.8281 \\ % w/ens
        %SEND-SupCSI ($r_{max}=-\tfrac{1}{4}$) & 0.8859 & 0.5941 & 0.9455 & 0.9107 & 0.9296 & 0.8750 & 0.9705 & 0.8474 & 0.8141 & 0.6650 & ?\\
        %EigenScore1 & 0.9969 & 0.9871 & 0.9867 & 0.9177 & 0.9866 & 0.9887 & 0.9890 & 0.9843 & 0.9796 \\
        \bottomrule
    \end{tabular}%
}}

\putTable{loo_pdpfa}{Leave-one-class-out detection rate (\%) at a fixed false alarm rate of 25\%}{
    \scalebox{0.73}{%
    \begin{tabular}{l c c c c c c c c c c c}
        \toprule
        &\multicolumn{10}{c}{CIFAR-10 Novelty Class} & \\
        \cmidrule(r){2-11}
        Method & Plane & Car & Bird & Cat & Deer & Dog & Frog & Horse & Ship & Truck & Avg.\\
        \midrule
        CSI \citeo{Tack:NIPS:20} & \bf 73.6 & 13.5 & \bf 89.8 & 87.8 & \bf 62.4 & \bf 68.4 & \bf 97.2 & 33.7 & \bf 45.6 & 30.8 & 60.3 \\ 
        %\multirow{2}{*}{No} & CSI \citep{Tack:NIPS:20} & 76.6 & 15.4 & \bf 90.4 & \bf 90.3 & 61.9 & \bf 66.8 & \bf 98.3 & 36.8 & \bf 50.4 & \bf 36.9 & \bf 62.4 \\  % w/ens
        %CSI \citep{Tack:NIPS:20} & ? & ? & ? & ? & ? & ? & ? & ? & ? & ? & ? \\  % w/ens no crop
        %SEND-CSI & 73.2 & 15.6 & 88.9 & 88.2 & 61.0 & 67.6 & 95.8 & 40.9 & 42.2 & 32.0 & 60.5 \\ % ONLY CSI Scores, w/out ens, eps=0.0
        glrt-CSI & 73.5 & \bf 15.6 & 89.0 & \bf 88.3 & 61.3 & 68.1 & 95.8 & \bf 41.1 & 42.7 & \bf 32.2 & \bf 60.7 \\ % ONLY CSI Scores, w/out ens, eps=0.25
                  %& SEND-CSI & 72.3 & 13.6 & 89.3 & 87.9 & 61.0 & 65.5 & 96.9 & 34.2 & 44.0 & 29.8 & 59.5 \\ % w/out ens, con + shift eps = 0.25
        %SEND-CSI & 76.9 & 16.0 & 90.3 & 89.6 & 62.4 & 64.0 & 97.5 & 40.4 & 45.9 & 35.6 & 61.9 \\ % ONLY CSI Scores, w/ ens, eps=0.0
                  %& SEND-CSI & \bf 76.8 & \bf 16.1 & \bf90.4 & 89.9 & \bf 62.6 & 64.6 & 97.7 & \bf 40.5 & 46.2 & 35.8 & 62.1 \\ % ONLY CSI Scores, w/ ens, eps=0.25
        \midrule
        %\multirow{4}{*}{Yes} & Mah \citep{Lee:NIPS:18} & ? & ? & ? & ? & ? & ? & ? & ? & ? & ? & ? \\ 
        Softmax \citeo{Hendrycks:ICLR:17} & 76.6 & 36.9 & 82.8 & 80.7 & 86.4 & 76.2 & 86.6 & 74.6 & 62.2 & 56.9 & 72.0 \\
        Maha \citeo{Lee:NIPS:18} & 57.3 & 46.1 & 54.1 & 65.6 & 29.5 & 46.8 & 20.6 & 62.8 & 45.9 & 49.3 & 47.8 \\
        %Mah \citep{Lee:NIPS:18} & 57.4 & 59.5 & 47.9 & 57.2 & 26.2 & 29.7 & 14.0 & 58.5 & 55.6 & 59.3 & 46.5 \\ 
        SupCSI \citeo{Tack:NIPS:20} & \bf 88.8 & 38.6 & \bf 96.6 & 89.7 & 93.9 & \bf 89.5 & 99.8 & 80.3 & 70.0 & 47.5 & 79.5 \\ 
                             %& NCIS-CSI & ? & ? & ? & ? & ? & ? & ? & ? & ? & ? & ? \\ 
                             %& SEND-CSI & ? & ? & ? & ? & ? & ? & ? & ? & ? & ? & ? \\ 
                             %& SEND-CSI & ? & ? & ? & ? & ? & ? & ? & ? & ? & ? & ? \\ 
                             %& SEND-CSI & ? & ? & ? & ? & ? & ? & ? & ? & ? & ? & ? \\ 
        %SEND-SupCSI & 89.3 & 39.6 & 95.6 & 90.5 & 94.9 & 87.0 & 99.9 & 81.2 & 76.0 & 51.6 & 80.6 \\ %eps=0
        glrt-SupCSI & 88.7 & \bf 38.8 & 96.5 & \bf 90.7 & \bf 95.3 & 87.7 & \bf 100. & \bf 81.8 & \bf 74.3 & \bf 48.7 & \bf 80.3 \\ %eps=1/4
                     %& Full-SEND & 85.6 & 37.0 & 93.2 & 90.9 & 82.8 & 79.7 & 98.9 & 69.1 & 55.2 & ? & ? \\ % w/out ens, CSI-mah dist
                     %& Full-SEND & 85.1 & 35.2 & 93.9 & 92.8 & 84.1 & ? & ? & ? & ? & 51.0 & ? \\ % w/out ens, CSI-mah dist, no ocsvm
                     %& Full-SEND & 81.7 & ? & ? & ? & ? & ? & ? & ? & ? & ? & ? \\ % w/out ens, CSI-mah dist, no ocsvm, no mah
        glrt-SupCSI+ & 84.4 & 36.3 & 93.4 & 92.1 & 81.9 & 81.0 & 98.9 & 74.2 & 56.5 & 53.1 & 75.2 \\ % w/out ens, full
                     %& Full-SEND & 82.3 & 34.7 & 93.8 & 93.0 & 82.4 & 80.7 & 99.2 & 71.1 & 58.6 & 49.6 & 74.5 \\ % w/out ens, joint-mah, no ocsvm
        \bottomrule
    \end{tabular}%
}}

Tables~\ref{tab:loo_auroc_scores} and \ref{tab:loo_pdpfa_scores} compare the proposed GLRT to several existing score-combining rules using the AUROC and DR metrics, respectively.
Both tables show the GLRT outperforming all other methods on average.
The other methods perform similarly, except for the ALR \citep{Walther:IMSC:13}, which performs noticeably worse.

\putTable{loo_base_scores}{Leave-one-class-out AUROC for different base scores}{
    \scalebox{0.73}{%
      \begin{tabular}{l c c c c c c c c c c c}
        \toprule
        &\multicolumn{10}{c}{CIFAR-10 Novelty Class} & \\
        \cmidrule(r){2-11}
        Method & Plane & Car & Bird & Cat & Deer & Dog & Frog & Horse & Ship & Truck & Avg.\\
        \midrule
            %Novelty Class & 0 & 1 & 2 & 3 & 4 & 5 & 6 & 7 & 8 & 9 & Avg.\\
        %\hline
        %Mah  & 0.9041 & 0.7175 & 0.6209 & 0.7077 & 0.9069 & 0.9122 & 0.7809 & 0.7692 & 0.7899 \\
        $\scorenormone$ & 0.7481 & 0.389 & 0.8308 & 0.8593 & 0.7423 & 0.7641 & 0.8266 & 0.6436 & 0.6873 & 0.5684 & 0.7060 \\
        $\scorecosone$ & 0.7705 & 0.4869 & 0.8479 & 0.8784 & 0.7341 & 0.7992 & 0.8267 & 0.7053 & 0.6366 & 0.5645 & 0.7250 \\
        $\scoreshiftone$ & 0.8259 & 0.4405 & 0.8835 & 0.8538 & 0.7302 & 0.7625 & 0.9245 & 0.5406 & 0.7007 & 0.5774 & 0.7240 \\
        $\scoresupocsvmone$ & 0.7216 & 0.5936 & 0.6418 & 0.6116 & 0.5451 & 0.5315 & 0.6629 & 0.5630 & 0.4702 & 0.6743 & 0.6016\\
        %OC-SVM & 0.9813 & 0.7581 & 0.6998 & 0.6891 & 0.9929 & 0.9921 & 0.9862 & 0.9774 & 0.8846 \\
        %Mah \citep{Lee:NIPS:18} & 0.7186 & \bf 0.7710 & 0.6900 & 0.7123 & 0.5336 & 0.6180 & 0.4795 & 0.7714 & 0.6692 & \bf 0.7950 & 0.6759 \\
        $\scoresupcsione$ & \bf 0.8575 & 0.5868 & \bf 0.9247 & \bf 0.8902 & \bf 0.9024 & \bf 0.8607 & \bf 0.9538 & \bf 0.8198 & \bf 0.7879 & 0.6564 & \bf 0.8240 \\
        %CSI \citep{Tack:NIPS:20} & 0.8230 & 0.4279 & 0.8989 & 0.8889 & 0.7582 & 0.7950 & 0.9298 & 0.5893 & 0.6983 & 0.5876 & 0.7397 \\
        %SupCSI \citep{Tack:NIPS:20} & 0.8827 & 0.5851 & 0.9437 & 0.9026 & 0.9231 & \bf 0.8793 & 0.9690 & \bf 0.8514 & 0.7913 & 0.6406 & 0.8369 \\
        %CSI-Mah & 0.7628 & 0.7729 & 0.7079 & 0.7072 & 0.5754 & 0.6118 & 0.4788 & 0.7328 & 0.6603 & 0.8172 & 0.6827 \\
        $\scoresupmahone$ & 0.7186 & \bf 0.7710 & 0.6900 & 0.7123 & 0.5336 & 0.6180 & 0.4795 & 0.7714 & 0.6692 & \bf 0.7948 & 0.6758 \\
        %END-4 & 0.9892 & 0.9817 & 0.9838 & 0.8479 & 0.9441 & 0.9493 & 0.9748 & 0.9643 \\
        %SEND-16 & 0.9950 & 0.9768 & 0.9787 & 0.8584 & 0.9888 & 0.9890 & 0.9801 & 0.9726 \\
        %SEND-24 ($z_{max}=0.0$)& 0.8656 & 0.6244 & 0.9218 & 0.9067 & 0.8407 & 0.8299 & 0.9414 & 0.8091 & 0.7251 & 0.7193 & ? \\ % z_min = 0.0
        %END-CSI & 0.8121 & 0.4606 & 0.8666 & 0.8695 & 0.7203 & 0.7698 & 0.8998 & 0.6581 & 0.6336 & 0.5984 & 0.7289 \\
        %SEND-CSI & 0.8230 & 0.4656 & 0.8870 & 0.8861 & 0.7379 & 0.7904 & 0.9147 & 0.6731 & 0.6341 & 0.6143 & 0.7426 \\
        %SEND-SupCSI & \bf 0.8866 & 0.5971 & \bf 0.9448 & 0.9100 & \bf 0.9299 & 0.8729 & \bf 0.9704 & 0.8474 & \bf 0.8186 & 0.6745 & \bf 0.8452\\ % z_max = 0
        %Full-SEND & 0.8768 & 0.6099 & 0.9270 & \bf 0.9121 & 0.8511 & 0.8427 & 0.9426 & 0.8151 & 0.7497 & 0.7121 & 0.8239 \\ % z_max = -0.25
        %SEND-SupCSI ($z_{max}=-\tfrac{1}{4}$) & 0.8859 & 0.5941 & 0.9455 & 0.9107 & 0.9296 & 0.8750 & 0.9705 & 0.8474 & 0.8141 & 0.6650 & ?\\
        %EigenScore1 & 0.9969 & 0.9871 & 0.9867 & 0.9177 & 0.9866 & 0.9887 & 0.9890 & 0.9843 & 0.9796 \\
        \bottomrule
    \end{tabular}%
}}

\putTable{loo_auroc_scores}{Leave-one-class-out AUROC of SupCSI+ with different score-combining techniques}{
    \scalebox{0.73}{%
        \begin{tabular}{l c c c c c c c c c c c}
        \toprule
        &\multicolumn{10}{c}{CIFAR-10 Novelty Class} & \\
        \cmidrule(r){2-11}
        Method & Plane & Car & Bird & Cat & Deer & Dog & Frog & Horse & Ship & Truck & Avg.\\
        \midrule
        %Novelty Class & 0 & 1 & 2 & 3 & 4 & 5 & 6 & 7 & 8 & 9 & Avg.\\
        %Full-SEND & 0.111, 0.414, 0.609, 0.793 & 0.007, 0.056, 0.13, 0.288 & 0.226, 0.582, 0.786, 0.909 & 0.174, 0.510, 0.714, 0.883 & 0.054, 0.267, 0.447, 0.723 & 0.039, 0.201, 0.376, 0.695 & 0.146, 0.576, 0.828, 0.969 & 0.018, 0.141, 0.309, 0.635 & 0.026, 0.147, 0.255, 0.461 & 0.018, 0.138, 0.229, 0.440 & ? \\
        %Full-SEND & 0.609 & 0.13 & \bf 0.786 & \bf 0.714 & 0.447 & 0.376 & \bf 0.828 & 0.309 & 0.255 & 0.229 & \bf 0.468 \\
        GLRT & 0.8768 & 0.6099 & \bf 0.9270 & 0.9121 & 0.8511 & 0.8427 & \bf 0.9426 & \bf 0.8151 & 0.7497 & 0.7120 & \bf 0.8239 \\
        Fisher & 0.8707 & 0.5990 & 0.9219 & 0.9112 & 0.8309 & 0.8374 & 0.9373 & 0.7972 & 0.7311 & 0.7105 & 0.8147 \\
        Bonferroni & 0.8513 & \bf 0.6413 & 0.9150 & 0.8760 & \bf 0.8688 & 0.7985 & 0.9383 & 0.8106 & 0.7376 & \bf 0.7191 & 0.8157 \\
        Simes/BH & 0.8559 & 0.6296 & 0.9189 & 0.8878 & 0.8670 & 0.8127 & 0.9418 & 0.8126 & 0.7367 & 0.7093 & 0.8172 \\
        ALR & 0.8664 & 0.5993 & 0.8323 & 0.8318 & 0.7264 & 0.7417 & 0.7837 & 0.6589 & 0.6746 & 0.7095 & 0.7425 \\
        Stouffer & \bf 0.8845 & 0.5846 & 0.9250 & \bf 0.9154 & 0.8387 & \bf 0.8507 & 0.9301 & 0.7930 & \bf 0.7575 & 0.6974 & 0.8177 \\
        \bottomrule
        \end{tabular}%
}}

% saved values for 1, 5, 10, and 20 %
\putTable{loo_pdpfa_scores}{Leave-one-class-out detection rate (\%) at a false-alarm rate of 25\% for SupCSI+ with different score-combining techniques}{
    \scalebox{0.73}{%
        \begin{tabular}{l c c c c c c c c c c c}
        \toprule
        &\multicolumn{10}{c}{CIFAR-10 Novelty Class} & \\
        \cmidrule(r){2-11}
        Method & Plane & Car & Bird & Cat & Deer & Dog & Frog & Horse & Ship & Truck & Avg.\\
        \midrule
        %Novelty Class & 0 & 1 & 2 & 3 & 4 & 5 & 6 & 7 & 8 & 9 & Avg.\\
        %\hline
        %Full-SEND & 0.111, 0.414, 0.609, 0.793 & 0.007, 0.056, 0.13, 0.288 & 0.226, 0.582, 0.786, 0.909 & 0.174, 0.510, 0.714, 0.883 & 0.054, 0.267, 0.447, 0.723 & 0.039, 0.201, 0.376, 0.695 & 0.146, 0.576, 0.828, 0.969 & 0.018, 0.141, 0.309, 0.635 & 0.026, 0.147, 0.255, 0.461 & 0.018, 0.138, 0.229, 0.440 & ? \\
        %Full-SEND & 0.609 & 0.13 & \bf 0.786 & \bf 0.714 & 0.447 & 0.376 & \bf 0.828 & 0.309 & 0.255 & 0.229 & \bf 0.468 \\
        GLRT & 84.4 & 36.3 & 93.4 & 92.1 & 81.9 & 81.0 & \bf 98.9 & \bf 74.2 & 56.5 & 53.1 & \bf 75.18 \\
        Fisher & 83.7 & 33.1 & 92.1 & 91.7 & 76.1 & 80.1 & 98.6 & 69.8 & 54.8 & 53.0 & 73.3 \\
        Bonferroni & 81.1 & 41.5 & 91.8 & 84.7 & 85.4 & 69.5 & 95.8 & 72.5 & 56.7 & 54.0 & 73.3 \\
        Simes/BH & 81.3 & \bf 42.1 &  92.7 & 87.0 & \bf 86.0 & 70.8 & 97.5 & 72.3 & \bf 58.1 & 55.1 & 74.3 \\
        ALR & 82.7 & 37.1 & 74.7 & 74.9 & 50.1 & 55.7 & 63.9 & 41.7 & 48.5 & \bf 60.4 & 59.0 \\
        Stouffer & \bf 86.0 & 33.0 & \bf 94.1 & \bf 92.8 & 77.0 & \bf 84.8 & 98.1 & 66.1 & 55.3 & 49.8 & 73.7 \\
        \bottomrule
        \end{tabular}%
}}

\section{Conclusion} \label{sec:conc}

State-of-the-art self-supervised OOD detectors generate several intermediate scores using pretext tasks, augmented views, and/or distribution shifts, and then heuristically combine them to form a final inlier score.
We proposed a principled approach to score combining based on null-hypothesis testing.
In particular, we posed the non-negative means problem \eqref{NMP}, solved it using a GLRT, and used that GLRT to combine the intermediate scores of the CSI and SupCSI techniques, as well as additional scores.
We also proposed an extension that uses a validation dataset to achieve probabilistic guarantees on the false-alarm rate.
Both dataset-vs-dataset experiments and leave-one-class-out experiments showed that our GLRT-based approach outperformed the state-of-the-art CSI and SupCSI methods \citep{Tack:NIPS:20} in both AUROC and detection rate for a fixed false-alarm rate.
In addition, our GLRT-based approach outperformed the traditional score-combining methods of Fisher, Bonferroni, Simes/BH, Stouffer, and ALR for our OOD detection tasks. 

\textbf{Limitations.}
One limitation of our approach is that, given a set of base scores, it is not clear which to combine.
Our experiments showed that, in some cases, incorporating additional scores reduced performance.
Another limitation of our approach is that it builds on self-supervised methods that rely on a set of transformations $\mc{T}$.
For images, good transformations are known, but for other modalities (e.g., audio or tabular data) it is not clear which transformations to use.

%==============================================================================
\acks{This work was supported in part by the National Science Foundation under grant 1539961 and the Air Force Research Laboratory under grant FA8650-32-C-1174. This work has been cleared for public release via PA Approval number AFRL-2023-2205.}
%==============================================================================

\clearpage
\bibliography{macros_abbrev,books,misc,machine,sparse,phase,kaggle_bib} % in /bib of group SVN

\begin{thebibliography}{58}
\providecommand{\natexlab}[1]{#1}
\providecommand{\url}[1]{\texttt{#1}}
\expandafter\ifx\csname urlstyle\endcsname\relax
  \providecommand{\doi}[1]{doi: #1}\else
  \providecommand{\doi}{doi: \begingroup \urlstyle{rm}\Url}\fi

\bibitem[Ahmed and Courville(2020)]{Ahmed:AAAI:20}
F.~Ahmed and A.~Courville.
\newblock Detecting semantic anomalies.
\newblock In \emph{Proc. AAAI Conf. Artificial Intell.}, volume~34, pages
  3154--3162, 2020.

\bibitem[Anderson et~al.(2012)Anderson, Dahl, and
  Vandenberghe]{Anderson:CVXOPT:12}
M.~S. Anderson, J.~Dahl, and L.~Vandenberghe.
\newblock {CVXOPT}: A python package for convex optimization, 2012.
\newblock URL \url{abel.ee.ucla.edu/cvxopt}.

\bibitem[Benjamini and Hochberg(1995)]{Benjamini:JRSSb:95}
Y.~Benjamini and Y.~Hochberg.
\newblock Controlling the false discovery rate: a practical and powerful
  approach to multiple testing.
\newblock \emph{J. Roy. Statist. Soc. B}, 57\penalty0 (1):\penalty0 289--300,
  1995.

\bibitem[Bergamin et~al.(2022)Bergamin, Mattei, Havtorn, Senetaire, Schmutz,
  Maal{\o}e, Hauberg, and Frellsen]{Bergamin:AISTATS:22}
F.~Bergamin, P.-A. Mattei, J.~D. Havtorn, H.~Senetaire, H.~Schmutz,
  L.~Maal{\o}e, S.~Hauberg, and J.~Frellsen.
\newblock Model-agnostic out-of-distribution detection using combined
  statistical tests.
\newblock In \emph{Proc. Intl. Conf. Artificial Intell. Statist.}, pages
  10753--10776, 2022.

\bibitem[Bergman and Hoshen(2020)]{Bergman:ICLR:20}
L.~Bergman and Y.~Hoshen.
\newblock Classification-based anomaly detection for general data.
\newblock In \emph{Proc. Intl. Conf. Learn. Rep.}, 2020.

\bibitem[Berk and Jones(1979)]{Berk:ZWVG:79}
R.~H. Berk and D.~H. Jones.
\newblock Goodness-of-fit test statistics that dominate the {K}olmogorov
  statistics.
\newblock \emph{Zeitschrift f{\"u}r Wahrscheinlichkeitstheorie und verwandte
  Gebiete}, 47\penalty0 (1):\penalty0 47--59, 1979.

\bibitem[Bishop(1994)]{Bishop:IEEV:94}
C.~M. Bishop.
\newblock Novelty detection and neural network validation.
\newblock \emph{IEE Proc. V: Vision, Image \& Signal Process.}, 141\penalty0
  (4):\penalty0 217--222, 1994.

\bibitem[Breunig et~al.(2000)Breunig, Kriegel, Ng, and Sander]{Breunig:ICMD:00}
M.~M. Breunig, H.-P. Kriegel, R.~T. Ng, and J.~Sander.
\newblock {LOF: I}dentifying density-based local outliers.
\newblock In \emph{Proc. ACM SIGMOD Intl. Conf. Manag. Data}, pages 93--104,
  2000.

\bibitem[Chalapathy and Chawla(2019)]{Chalapathy:19}
R.~Chalapathy and S.~Chawla.
\newblock Deep learning for anomaly detection: {A} survey.
\newblock \emph{arXiv:1901.03407}, 2019.

\bibitem[Chen et~al.(2020)Chen, Kornblith, Norouzi, and Hinton]{Chen:ICML:20}
T.~Chen, S.~Kornblith, M.~Norouzi, and G.~Hinton.
\newblock A simple framework for contrastive learning of visual
  representations.
\newblock In \emph{Proc. Intl. Conf. Mach. Learn.}, pages 1597--1607, 2020.

\bibitem[Cousins(2007)]{Cousins:07}
R.~D. Cousins.
\newblock Annotated bibliography of some papers on combining significances or
  p-values.
\newblock \emph{arXiv:0705.2209}, 2007.

\bibitem[Deecke et~al.(2021)Deecke, Ruff, Vandermeulen, and
  Bilen]{Deecke:ICML:21}
L.~Deecke, L.~Ruff, R.~Vandermeulen, and H.~Bilen.
\newblock Transfer-based semantic anomaly detection.
\newblock In \emph{Proc. Intl. Conf. Mach. Learn.}, pages 2546--2558, 2021.

\bibitem[Deng et~al.(2009)Deng, Dong, Socher, Li, Li, and
  Fei-Fei]{Deng:CVPR:09}
J.~Deng, W.~Dong, R.~Socher, L.-J. Li, K.~Li, and L.~Fei-Fei.
\newblock Imagenet: {A} large-scale hierarchical image database.
\newblock In \emph{Proc. IEEE Conf. Comp. Vision Pattern Recog.}, pages
  248--255, 2009.

\bibitem[Di~Mattia et~al.(2019)Di~Mattia, Galeone, De~Simoni, and
  Ghelfi]{DiMattia:19}
F.~Di~Mattia, P.~Galeone, M.~De~Simoni, and E.~Ghelfi.
\newblock A survey on {GANs} for anomaly detection.
\newblock \emph{arXiv:1906.11632}, 2019.

\bibitem[Donoho and Jin(2004)]{Donoho:AS:04}
D.~Donoho and J.~Jin.
\newblock Higher criticism for detecting sparse heterogeneous mixtures.
\newblock \emph{Ann. Statist.}, 32\penalty0 (3):\penalty0 962--994, 2004.

\bibitem[Ericsson et~al.(2022)Ericsson, Gouk, Loy, and
  Hospedales]{Ericsson:SPM:22}
L.~Ericsson, H.~Gouk, C.~C. Loy, and T.~M. Hospedales.
\newblock Self-supervised representation learning: {I}ntroduction, advances,
  and challenges.
\newblock \emph{IEEE Signal Process. Mag.}, 39\penalty0 (3):\penalty0 42--62,
  2022.

\bibitem[Fisher(1992)]{Fisher:Book:92}
R.~A. Fisher.
\newblock \emph{Statistical methods for research workers}.
\newblock Springer, 1992.

\bibitem[Georgescu et~al.(2021)Georgescu, Barbalau, Ionescu, Khan, Popescu, and
  Shah]{Georgescu:CVPR:21}
M.-I. Georgescu, A.~Barbalau, R.~T. Ionescu, F.~S. Khan, M.~Popescu, and
  M.~Shah.
\newblock Anomaly detection in video via self-supervised and multi-task
  learning.
\newblock In \emph{Proc. IEEE Conf. Comp. Vision Pattern Recog.}, pages
  12742--12752, 2021.

\bibitem[Gidaris et~al.(2018)Gidaris, Singh, and Komodakis]{Gidaris:ICLR:18}
S.~Gidaris, P.~Singh, and N.~Komodakis.
\newblock Unsupervised representation learning by predicting image rotations.
\newblock In \emph{Proc. Intl. Conf. Learn. Rep.}, Apr. 2018.

\bibitem[Golan and El-Yaniv(2018)]{Golan:NIPS:18}
I.~Golan and R.~El-Yaniv.
\newblock Deep anomaly detection using geometric transformations.
\newblock In \emph{Proc. Neural Info. Process. Syst. Conf.}, volume~31, 2018.

\bibitem[Haroush et~al.(2022)Haroush, Frostig, Heller, and
  Soudry]{Haroush:ICLR:22}
M.~Haroush, T.~Frostig, R.~Heller, and D.~Soudry.
\newblock A statistical framework for efficient out of distribution detection
  in deep neural networks.
\newblock In \emph{Proc. Intl. Conf. Learn. Rep.}, 2022.

\bibitem[He et~al.(2016)He, Zhang, Ren, and Sun]{He:CVPR:16}
K.~He, X.~Zhang, S.~Ren, and J.~Sun.
\newblock Deep residual learning for image recognition.
\newblock In \emph{Proc. IEEE Conf. Comp. Vision Pattern Recog.}, pages
  770--778, 2016.

\bibitem[Hendrycks and Gimpel(2017)]{Hendrycks:ICLR:17}
D.~Hendrycks and K.~Gimpel.
\newblock A baseline for detecting misclassified and out-of-distribution
  examples in neural networks.
\newblock In \emph{Proc. Intl. Conf. Learn. Rep.}, 2017.

\bibitem[Hendrycks et~al.(2019{\natexlab{a}})Hendrycks, Mazeika, and
  Dietterich]{Hendrycks:ICLR:19}
D.~Hendrycks, M.~Mazeika, and T.~Dietterich.
\newblock Deep anomaly detection with outlier exposure.
\newblock In \emph{Proc. Intl. Conf. Learn. Rep.}, 2019{\natexlab{a}}.

\bibitem[Hendrycks et~al.(2019{\natexlab{b}})Hendrycks, Mazeika, Kadavath, and
  Song]{Hendrycks:NIPS:19}
D.~Hendrycks, M.~Mazeika, S.~Kadavath, and D.~Song.
\newblock Using self-supervised learning can improve robustness and
  uncertainty.
\newblock In \emph{Proc. Neural Info. Process. Syst. Conf.},
  2019{\natexlab{b}}.

\bibitem[Hoffmann(2007)]{Hoffmann:PR:07}
H.~Hoffmann.
\newblock Kernel {PCA} for novelty detection.
\newblock \emph{Pattern Recognition}, 40\penalty0 (3):\penalty0 863--874, 2007.

\bibitem[Kaur et~al.(2022)Kaur, Jha, Roy, Park, Dobriban, Sokolsky, and
  Lee]{Kaur:AAAI:22}
R.~Kaur, S.~Jha, A.~Roy, S.~Park, E.~Dobriban, O.~Sokolsky, and I.~Lee.
\newblock {iDECODe: I}n-distribution equivariance for conformal
  out-of-distribution detection.
\newblock In \emph{Proc. AAAI Conf. Artificial Intell.}, volume~36, pages
  7104--7114, 2022.

\bibitem[Khalid et~al.(2022)Khalid, Esmaeili, Karim, and
  Rahnavard]{Khalid:CVPRW:22}
U.~Khalid, A.~Esmaeili, N.~Karim, and N.~Rahnavard.
\newblock {RODD: A} self-supervised approach for robust out-of-distribution
  detection.
\newblock In \emph{Proc. IEEE Conf. Comp. Vision Pattern Recog. Workshop},
  pages 163--170, 2022.

\bibitem[Khosla et~al.(2020)Khosla, Teterwak, Wang, Sarna, Tian, Isola,
  Maschinot, Liu, and Krishnan]{Khosla:NIPS:20}
P.~Khosla, P.~Teterwak, C.~Wang, A.~Sarna, Y.~Tian, P.~Isola, A.~Maschinot,
  C.~Liu, and D.~Krishnan.
\newblock Supervised contrastive learning.
\newblock In \emph{Proc. Neural Info. Process. Syst. Conf.}, volume~33, pages
  18661--18673, 2020.

\bibitem[Krizhevsky(2009)]{Krizhevsky:CIFAR:09}
A.~Krizhevsky.
\newblock Learning multiple layers of features from tiny images.
\newblock \emph{Technical Report}, 2009.
\newblock URL
  \url{https://www.cs.toronto.edu/~kriz/learning-features-2009-TR.pdf}.

\bibitem[Latecki et~al.(2007)Latecki, Lazarevic, and Pokrajac]{Latecki:MLDM:07}
L.~J. Latecki, A.~Lazarevic, and D.~Pokrajac.
\newblock Outlier detection with kernel density functions.
\newblock In \emph{Proc. Mach. Learn. Data Mining Pattern Recog.}, volume~7,
  pages 61--75, 2007.

\bibitem[Laurikkala et~al.(2000)Laurikkala, Juhola, Kentala, Lavrac, Miksch,
  and Kavsek]{Laurikkala:IDAMP:00}
J.~Laurikkala, M.~Juhola, E.~Kentala, N.~Lavrac, S.~Miksch, and B.~Kavsek.
\newblock Informal identification of outliers in medical data.
\newblock In \emph{Intl. Workshop Intell. Data Anal. Med. Pharma.}, pages
  20--24, 2000.

\bibitem[Lee et~al.(2018)Lee, Lee, Lee, and Shin]{Lee:NIPS:18}
K.~Lee, K.~Lee, H.~Lee, and J.~Shin.
\newblock A simple unified framework for detecting out-of-distribution samples
  and adversarial attacks.
\newblock In \emph{Proc. Neural Info. Process. Syst. Conf.}, 2018.

\bibitem[Lehmann et~al.(2005)Lehmann, Romano, and Casella]{Lehmann:Book:05}
E.~L. Lehmann, J.~P. Romano, and G.~Casella.
\newblock \emph{Testing Statistical Hypotheses}, volume~3.
\newblock Springer, 2005.

\bibitem[Magesh et~al.(2022)Magesh, Veeravalli, Roy, and Jha]{Magesh:22}
A.~Magesh, V.~V. Veeravalli, A.~Roy, and S.~Jha.
\newblock Multiple testing framework for out-of-distribution detection.
\newblock \emph{arXiv:2206.09522}, 2022.

\bibitem[Netzer et~al.(2011)Netzer, Wang, Coates, Bissacco, Wu, and
  Ng]{Netzer:NIPS:11}
Y.~Netzer, T.~Wang, A.~Coates, A.~Bissacco, B.~Wu, and A.~Y. Ng.
\newblock Reading digits in natural images with unsupervised feature learning.
\newblock In \emph{Proc. Neural Info. Process. Syst. Conf.}, 2011.

\bibitem[Pang et~al.(2021)Pang, Shen, Cao, and Hengel]{Pang:CS:21}
G.~Pang, C.~Shen, L.~Cao, and A.~V.~D. Hengel.
\newblock Deep learning for anomaly detection: {A} review.
\newblock \emph{ACM Comput. Surveys}, 54\penalty0 (2):\penalty0 1--38, 2021.

\bibitem[Ramaswamy et~al.(2000)Ramaswamy, Rastogi, and Shim]{Ramaswamy:ICMD:00}
S.~Ramaswamy, R.~Rastogi, and K.~Shim.
\newblock Efficient algorithms for mining outliers from large data sets.
\newblock In \emph{Proc. ACM SIGMOD Intl. Conf. Manag. Data}, pages 427--438,
  2000.

\bibitem[Reiss and Hoshen(2023)]{Reiss:AAAI:23}
T.~Reiss and Y.~Hoshen.
\newblock Mean-shifted contrastive loss for anomaly detection.
\newblock In \emph{Proc. AAAI Conf. Artificial Intell.}, Washington, DC, Feb.
  2023.

\bibitem[Reiss et~al.(2021)Reiss, Cohen, Bergman, and Hoshen]{Reiss:CVPR:21}
T.~Reiss, N.~Cohen, L.~Bergman, and Y.~Hoshen.
\newblock {PANDA: A}dapting pretrained features for anomaly detection and
  segmentation.
\newblock In \emph{Proc. IEEE Conf. Comp. Vision Pattern Recog.}, page
  2806–2814, virtual, June 2021.

\bibitem[Ruff et~al.(2020)Ruff, Vandermeulen, G\"{o}rnitz, Binder, M\"{u}ller,
  M\"{u}ller, and Kloft]{Ruff:ICLR:20}
L.~Ruff, R.~Vandermeulen, N.~G\"{o}rnitz, A.~Binder, E.~M\"{u}ller,
  K.~M\"{u}ller, and M.~Kloft.
\newblock Transfer-based semantic anomaly detection.
\newblock In \emph{Proc. Intl. Conf. Learn. Rep.}, 2020.

\bibitem[Ruff et~al.(2021)Ruff, Kauffmann, and Vandermeulen]{Ruff:PROC:21}
L.~Ruff, J.~R. Kauffmann, and R.~A. Vandermeulen.
\newblock A unifying review of deep and shallow anomaly detection.
\newblock \emph{Proc. IEEE}, 109\penalty0 (5):\penalty0 756--795, 2021.

\bibitem[Sch\"{o}lkopf et~al.(2001)Sch\"{o}lkopf, Platt, Shawe-Taylor, Smola,
  and Williamson]{Scholkopf:NC:01}
B.~Sch\"{o}lkopf, J.~C. Platt, J.~C. Shawe-Taylor, A.~J. Smola, and R.~C.
  Williamson.
\newblock Estimating the support of a high-dimensional distribution.
\newblock \emph{Neural Comput.}, 13\penalty0 (7):\penalty0 1443--1471, July
  2001.

\bibitem[Simes(1986)]{Simes:BIO:86}
R.~J. Simes.
\newblock An improved {B}onferroni procedure for multiple tests of
  significance.
\newblock \emph{Biometrika}, 73\penalty0 (3):\penalty0 751--754, 1986.

\bibitem[Smith and Topin(2019)]{Smith:SPIE:19}
L.~N. Smith and N.~Topin.
\newblock Super-convergence: {V}ery fast training of neural networks using
  large learning rate.
\newblock In T.~Pham, editor, \emph{Artificial Intell. Machine Learning for
  Multi-Domain Operations Applications}, volume 11006, page 11006 12. SPIE,
  2019.

\bibitem[Sohn et~al.(2021)Sohn, Li, Yoon, Jin, and Pfister]{Sohn:ICLR:21}
K.~Sohn, C.-L. Li, J.~Yoon, M.~Jin, and T.~Pfister.
\newblock Learning and evaluating representations for deep one-class
  classification.
\newblock In \emph{Proc. Intl. Conf. Learn. Rep.}, 2021.

\bibitem[Szegedy et~al.(2016)Szegedy, Vanhoucke, Ioffe, Shlens, and
  Wojna]{Szegedy:CVPR:16}
C.~Szegedy, V.~Vanhoucke, S.~Ioffe, J.~Shlens, and Z.~Wojna.
\newblock Rethinking the inception architecture for computer vision.
\newblock In \emph{Proc. IEEE Conf. Comp. Vision Pattern Recog.}, 2016.

\bibitem[Tack et~al.(2020)Tack, Mo, Jeong, and Shin]{Tack:NIPS:20}
J.~Tack, S.~Mo, J.~Jeong, and J.~Shin.
\newblock {CSI: N}ovelty detection via contrastive learning on distributionally
  shifted instances.
\newblock In \emph{Proc. Neural Info. Process. Syst. Conf.}, pages
  11839--11852, 2020.

\bibitem[Tax and Duin(2004)]{Tax:ML:04}
D.~Tax and R.~Duin.
\newblock Support vector data description.
\newblock \emph{Mach. Learn.}, 54:\penalty0 45--66, 2004.

\bibitem[Vovk(2012)]{Vovk:ACML:12}
V.~Vovk.
\newblock Conditional validity of inductive conformal predictors.
\newblock In \emph{Asian Conf. Mach. Learn.}, pages 475--490, 2012.

\bibitem[Walther(2013)]{Walther:IMSC:13}
G.~Walther.
\newblock The average likelihood ratio for large-scale multiple testing and
  detecting sparse mixtures.
\newblock \emph{Inst. Math. Stat.(IMS) Collect}, 9:\penalty0 317--326, 2013.

\bibitem[Wang and Isola(2020)]{Wang:ICML:20}
T.~Wang and P.~Isola.
\newblock Understanding contrastive representation learning through alignment
  and uniformity on the hypersphere.
\newblock In \emph{Proc. Intl. Conf. Mach. Learn.}, pages 9929--9939, 2020.

\bibitem[Wasserman(2004)]{Wasserman:Book:04}
L.~Wasserman.
\newblock \emph{All of Statistics: A Concise Course in Statistical Inference}.
\newblock Springer, 2004.

\bibitem[Wei et~al.(2019)Wei, Wainwright, and Guntuboyina]{Wei:AS:19}
Y.~Wei, M.~J. Wainwright, and A.~Guntuboyina.
\newblock The geometry of hypothesis testing over convex cones: {G}eneralized
  likelihood ratio tests and minimax radii.
\newblock \emph{Ann. Statist.}, 47\penalty0 (2):\penalty0 994--1024, 2019.

\bibitem[Winkens et~al.(2020)Winkens, Bunel, Roy, Stanforth, Natarajan, Ledsam,
  MacWilliams, Kohli, Karthikesalingam, Kohl, Cemgil, Eslami, and
  Ronneberger]{Winkens:20}
J.~Winkens, R.~Bunel, A.~G. Roy, R.~Stanforth, V.~Natarajan, J.~R. Ledsam,
  P.~MacWilliams, P.~Kohli, A.~Karthikesalingam, S.~Kohl, T.~Cemgil, S.~M.~A.
  Eslami, and O.~Ronneberger.
\newblock Contrastive training for improved out-of-distribution detection.
\newblock \emph{arXiv:2007.05566v1}, July 2020.

\bibitem[You et~al.(2017)You, Ginsburg, and Gitman]{You:17}
Y.~You, B.~Ginsburg, and I.~Gitman.
\newblock Large batch training of convolutional networks.
\newblock \emph{arXiv:1708:03888v3}, 2017.

\bibitem[Yu et~al.(2015)Yu, Seff, Zhang, Song, Funkhouser, and
  Xiao]{Yu:LSUN:15}
F.~Yu, A.~Seff, Y.~Zhang, S.~Song, T.~Funkhouser, and J.~Xiao.
\newblock {LSUN}: {C}onstruction of a large-scale image dataset using deep
  learning with humans in the loop.
\newblock \emph{arXiv:1506.03365}, 2015.

\bibitem[Zimek et~al.(2012)Zimek, Schubert, and Kriegel]{Zimek:SADM:12}
A.~Zimek, E.~Schubert, and H.-P. Kriegel.
\newblock A survey on unsupervised outlier detection in high-dimensional
  numerical data.
\newblock \emph{Stat. Anal. and Data Mining}, 5\penalty0 (5):\penalty0
  363--387, 2012.

\end{thebibliography}

%%%%%%%%%%%%%%%%%%%%%%%%%%%%%%%%%%%%%%%%%%%%%%%%%%%%%%%%%%%%
\clearpage
\appendix
%\section{Supplementary Material}

\section{Alternative formulations} \label{app:alternatives}

As discussed in \secref{glrt}, the NM problem \eqref{NMP} models the z-values as independent, even though we don't expect them to be independent in practice.
In this appendix, we discuss two alternative formulations that address this independence issue and explain their limitations.

\subsection{Multiple hypothesis testing} 
%\textbf{Multiple hypothesis testing.} 
One way to avoid the independence assumption in \eqref{NMP} is to instead use a \term{multiple hypothesis testing} formulation \citep{Wasserman:Book:04}, i.e., 
\begin{align}
\text{~for~} l=1,\dots,m,~~
\begin{cases}
H_{0l}:~Z_l \sim \mc{N}(0,1) \\
H_{1l}:~Z_l \sim \mc{N}(\mu_l,1) \text{~for some~} \mu_l\leq-\epsilon,
\end{cases}
\label{eq:NMPmulti}
\end{align}
from which the global inlier hypothesis (that $\vec{x}\sim\inlierPx$) would be constructed as $H_0=\cap_{l=1}^m H_{0l}$.
But, as we now show, \eqref{NMPmulti} reduces to the classical methods of score combining reviewed in \secref{hypo}.
To see why, recall that the test ``$z_l\leq\tau$'' is UMP for rejecting $H_{0l}$.
To combine $\{z_l\}_{l=1}^m$, one could use Stouffer's method or convert them to p-values $\bar{q}_l=\Phi(z_l)$ and use a classical p-value combining method like Bonferroni.
But since the transformations from the raw score $s_l(\vec{x})$ to $z_l$ to $\bar{q}_l$ are all strictly monotonically increasing, the p-value $\bar{q}_l$ will be identical to the p-value $q_l$ computed from the raw score $s_l(\vec{x})$, and so \eqref{NMPmulti} reduces to classical score-combining.
We numerically compare these classical score-combining approaches to the proposed GLRT in \secref{experiments}.

For completeness, we note that the $\epsilon=0$ case of \eqref{NMPmulti} was briefly discussed by Donoho and Jin in \citep{Donoho:AS:04}. 
For tractability, however, the authors approximated $H_{1l}$ by a Gaussian-mixture model and analyzed it in the sparse regime, which led them to propose the \term{higher criticism} (HC) statistic.
Later, Walther \citep{Walther:IMSC:13} compared the HC statistic to the \term{Berk-Jones} (BJ) statistic \citep{Berk:ZWVG:79} and found that one or the other may dominate, depending on the sparsity of the mixture.
Walther then proposed the \term{adjusted likelihood ratio} (ALR) statistic and empirically demonstrated that it performs similarly to the best of HC and BJ. 
We numerically investigate the ALR approach in \secref{experiments}.

\subsection{The negative-means problem with general covariances} 
%\textbf{The negative-means problem with general covariances.} 
Another way to avoid the independence assumption in \eqref{NMP} is to explicitly model correlation among the z-values via 
\begin{subequations}
\label{eq:NMPsigma2}
\begin{align}
&H_0:~\vec{Z} \sim \mc{N}(\vec{0},\vec{\Sigma}_0) \\
&H_1:~\vec{Z} \sim \mc{N}(\vec{\mu},\vec{\Sigma}_1) \text{~with~} \mu_l\leq-\epsilon \text{~for all~}l=1,\dots,m.
\end{align}
\end{subequations}
Although $\vec{\Sigma}_0$ could be chosen as the inlier sample-covariance matrix $\hvec{\Sigma}$, it's not clear how to choose $\vec{\Sigma}_1$ since we have no information about the novelties. 
One possibility is to use the inlier sample-covariance matrix for both $\vec{\Sigma}_0$ and $\vec{\Sigma}_1$.
 
In any case, we will study \eqref{NMPsigma2} further under the assumption that $\vec{\Sigma}_0=\vec{\Sigma}_1=\vec{\Sigma}$ for some $\vec{\Sigma}$. 
If $\vec{\mu}$ was known, then the log-LR (recall \eqref{LR}) would be  
\begin{align}
\log \LR 
&= -\tfrac{1}{2}\vec{z}\tran\vec{\Sigma}^{-1}\vec{z} + \tfrac{1}{2}(\vec{z}-\vec{\mu})\tran\vec{\Sigma}^{-1}(\vec{z}-\vec{\mu}) 
\label{eq:logLR_sigma} .
\end{align}
Since $\vec{\mu}$ is unknown, we could pose the GLRT
\begin{align}
\GLR(\vec{z}) \underset{H_1}{\overset{H_0}{\gtrless}} \tau 
\quad\text{for}\quad
\GLR(\vec{z}) = \frac{\mc{N}(\vec{z};\vec{0},\vec{\Sigma})}{\max_{\vec{\mu}: \mu_l \leq -\epsilon~\forall l} \mc{N}(\vec{z};\vec{\mu},\vec{\Sigma})},
\label{eq:glrt_sigma}
\end{align}
whose denominator is maximized by 
\begin{align}
\vec{\mu}^* 
&\defn \argmax_{\vec{\mu}: \mu_l \leq -\epsilon~\forall l} \mc{N}(\vec{z};\vec{\mu},\vec{\Sigma})
= \argmin_{\vec{\mu}: \mu_l \leq -\epsilon~\forall l} (\vec{z} - \vec{\mu})\tran \vec{\Sigma}^{-1}(\vec{z} - \vec{\mu}) 
\label{eq:z_minus_sigma}.
\end{align}
Since \eqref{z_minus_sigma} is a quadratic program, $\vec{\mu}^*$ could be readily computed using an optimization package like CVX \citep{Anderson:CVXOPT:12}.
Plugging $\vec{\mu}^*$ into the log-LR expression \eqref{logLR_sigma}, we obtain 
\begin{align}
\ln \GLR(\vec{z};\vec{\Sigma}) 
= (\tfrac{1}{2}\vec{\mu}^*(\vec{z})-\vec{z})\tran\vec{\Sigma}^{-1}\vec{\mu}^*(\vec{z}) 
\label{eq:logGLR_sigma} ,
\end{align}
which could be used as a test statistic.
In particular, we would reject the null hypothesis $H_0$ whenever $\ln\GLR(\vec{z};\vec{\Sigma})\geq\tau$ for an appropriately chosen threshold $\tau$.
In \eqref{logGLR_sigma}, we wrote ``$\vec{\mu}^*(\vec{z})$'' to emphasize the dependence of $\vec{\mu}^*$ on $\vec{z}$.

Recall that the proposed $\scoreglrt(\vec{z})$ in \eqref{logGLR} results from using $\ln \GLR(\vec{z};\vec{\Sigma})$ with $\vec{\Sigma}=\vec{I}$.
Another choice for $\vec{\Sigma}$ would be the inlier covariance, which could be computed from inlier training data $\mc{X}\train=\{\vec{x}_i\}$ via
\begin{align}
\hvec{\Sigma} 
\defn \frac{1}{n-1} \sum_{i=1}^n \vec{z}_i\vec{z}_i\tran + \sigma \vec{I}
\label{eq:sample_cov},
\end{align}
where $\vec{z}_i$ contains the empirical z-values associated with the training sample $\vec{x}_i$ and $\sigma>0$ is a small constant that ensures $\hvec{\Sigma}$ is full rank.
We use $\sigma=10^{-6}$ in our experiments.

We now empirically study the choice of $\vec{\Sigma}$ in $\ln \GLR(\vec{z};\vec{\Sigma})$ by repeating the dataset-vs-dataset experiments from \secref{experiments}.
\tabref{cov_selection_exp} shows the performance of SupCSI+ with $\ln \GLR(\vec{z};\vec{\Sigma})$-based score combining using $\vec{\Sigma}=\vec{I}$ versus $\vec{\Sigma}=\hvec{\Sigma}$.
There we see that $\vec{\Sigma}=\vec{I}$ leads to better AUROC performance than $\vec{\Sigma}=\hvec{\Sigma}$ in most experiments, and on average across experiments.

\putTable{cov_selection_exp}{Dataset-vs-dataset AUROC of SupCSI+ with $\ln\GLR(\vec{z};\vec{\Sigma})$-based score combining for two choices of $\vec{\Sigma}$}{

  \scalebox{0.73}{%
    \begin{tabular}{l c c c c c}
        \multicolumn{6}{c}{(a) Inlier dataset - CIFAR-10} \\
        \toprule
        & \multicolumn{4}{c}{novelty dataset} & \\
        \cmidrule(r){2-5}
        $\vec{\Sigma}$ & SVHN & LSUN & ImageNet & CIFAR-100 & Avg. \\
        \hline
        $\hvec{\Sigma}$ & 0.9712 & 0.9608 & 0.9683 & 0.8239 & 0.9311 \\ % sigma=1e-6
        %$\vec{I}$ & \bf 0.9950 & \bf 0.9768 & \bf 0.9787 & \bf 0.8584 & \bf 0.9522\\ INCONSISTENT WITH dvd_auroc TABLE 
        $\vec{I}$ & \bf 0.9968 & \bf 0.9836 & \bf 0.9840 & \bf 0.9064 & \bf 0.9677 \\
        \bottomrule
    \end{tabular}%
  \qquad
    \begin{tabular}{l c c c c c}
        \multicolumn{6}{c}{(b) Inlier dataset - SVHN} \\
        \toprule
        & \multicolumn{4}{c}{novelty dataset} & \\
        \cmidrule(r){2-5}
        $\vec{\Sigma}$ & SVHN & LSUN & ImageNet & CIFAR-100 & Avg. \\
        \hline
        $\hvec{\Sigma}$ & 0.9922 & 0.9917 & 0.9692 & 0.9618 & 0.9787 \\ % sigma=1e-6
        %$\vec{I}$ & 0.9888 & 0.9890 & \bf 0.9801 & \bf 0.9726 & \bf 0.9826 \\ INCONSISTENT WITH dvd_auroc TABLE
        $\vec{I}$ & \bf 0.9945 & \bf 0.9952 & \bf 0.9927 & \bf 0.9887 & \bf 0.9928 \\
        \bottomrule
    \end{tabular}%
  }
}

Next, we perform an eigen-analysis of the log-GLRT score in an attempt to better understand the reason that $\vec{\Sigma}=\vec{I}$ performs better than $\vec{\Sigma}=\hvec{\Sigma}$.
We write the eigen-decomposition of the inlier covariance matrix as 
\begin{align}
%\hvec{\Sigma} &= \vec{V \Lambda V}\tran 
\hvec{\Sigma} &= \sum_{k=1}^m \lambda_ k \vec{v}_k \vec{v}_k\tran 
\label{eq:cov_eig_decomp},
\end{align}
where $\{\vec{v}_k\}_{k=1}^n$ are orthonormal eigenvectors and $\{\lambda_k\}_{k=1}^m$ are non-negative eigenvalues.
The log-GLR from \eqref{logGLR_sigma} can then be written as
\begin{align}
\ln \GLR(\vec{z};\hvec{\Sigma}) 
&= \sum_{k=1}^m \frac{\scoreeigk(\vec{z})}{\lambda_k} 
\text{~~and~~}
\ln \GLR(\vec{z};\vec{I}) 
= \sum_{k=1}^m \scoreeigk(\vec{z}) 
\label{eq:glrt_eig_decomp}\\
\text{for~~}
\scoreeigk(\vec{z}) 
&\defn (\tfrac{1}{2}\vec{\mu}^*(\vec{z}) - \vec{z})\tran\vec{v}_k \cdot \vec{v}_k\tran \vec{\mu}^*(\vec{z})
\label{eq:scoreeigk} ,
\end{align}
%[Note: This used to be $\scoreeigk(\vec{z})=\vec{v}_k\tran \vec{z}$, in which case it's hard to justify the results of the eigen experiment]
where the second equality in \eqref{glrt_eig_decomp} follows from the fact that $\sum_{k=1}^m \vec{v}_k\vec{v}_k\tran=\vec{I}$.

\begin{figure}
    \centering
    \includegraphics[width=0.49\textwidth]{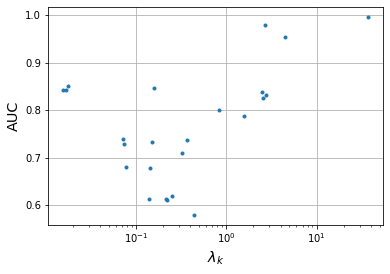}%
    \hfill
    \includegraphics[width=0.49\textwidth]{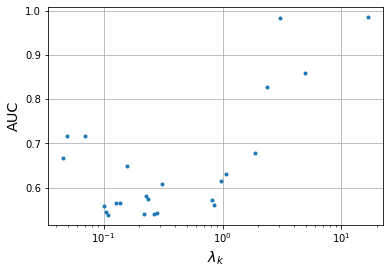}
    \vspace{2mm}
    \caption{AUROC of eigen-score $\scoreeigk(\vec{z})$ versus eigenvalue $\lambda_k$ for $k=1,\dots,m$ on two $m=24$ score experiments from \secref{experiments}: CIFAR-10 inliers with SVHN novelties (left) and SVHN novelties with CIFAR-10 inliers (right).%  The least-squares regression line and $R^2$ coefficient-of-determination are also superimposed. 
}
\vspace{-2mm}
\label{fig:eigen_score}
\end{figure}

In \figref{eigen_score}, we plot the AUROC performance of the ``eigen-score'' $\scoreeigk$ versus the eigenvalue $\lambda_k$ for the two experiments from \tabref{cov_selection_exp}.
%Due to the inherent sign ambiguity of the eigen-vectors, $\vec{v}_k$, we will negate the eigen-score if its AUROC is less than $0.5$.
Both experiments show a positive correlation between AUROC performance of the eigen-score $\scoreeigk(\vec{z})$ and the size of the corresponding eigenvalue $\lambda_k$.
Thus, dividing $\scoreeigk(\vec{z})$ by $\lambda_k$, as done when constructing $\ln\GLR(\vec{z};\hvec{\Sigma})$, will devalue the better-performing tests.
Conversely, not dividing $\scoreeigk(\vec{z})$ by $\lambda_k$, as done when constructing $\ln\GLR(\vec{z};\vec{I})$, will lead to better performance.

\end{document}